\pdfoutput=1
\documentclass[11pt]{article}

\usepackage{acl}

\usepackage{times}
\usepackage{latexsym}

\usepackage[T1]{fontenc}
\usepackage[utf8]{inputenc}

\usepackage{microtype}

\usepackage{inconsolata}

\usepackage{graphicx}

\usepackage{amsmath}
\usepackage{amssymb}
\usepackage{booktabs}
\usepackage{dsfont}
\usepackage{hyperref}
\usepackage{multirow}
\usepackage{listings}
\usepackage{subcaption}
\usepackage{tabularx}
\usepackage{tcolorbox}

\DeclareMathOperator*{\expectation}{\mathbb{E}}
\DeclareMathOperator*{\mean}{mean}
\DeclareMathOperator*{\topk}{topk}

\newcommand{\cm}{\checkmark}
\newcommand{\tb}{\textbf}
\newcommand{\ul}{\underline}

\newcommand{\ra}[1]{\renewcommand{\arraystretch}{#1}}

\lstdefinestyle{prompt}{
    basicstyle=\ttfamily\small,
    frame=single,
    breaklines=true,
    backgroundcolor=\color{gray!10}
}

\title{
    ConSim: Measuring Concept-Based Explanations'\\
    Effectiveness with Automated Simulatability
}

\renewcommand\paragraph[1]{
    \vspace{0.2cm}
    \noindent 
    \textbf{#1}
}

\author{
 \textbf{Antonin Poché\textsuperscript{1,2}},
 \textbf{Alon Jacovi\textsuperscript{3}},\\
 \textbf{Agustin Martin Picard\textsuperscript{1}},
 \textbf{Victor Boutin\textsuperscript{4,5}},\\
 \textbf{Fanny Jourdan\textsuperscript{1}},
\\
\\
 \textsuperscript{1}IRT Saint Exupéry,
 \textsuperscript{2}IRIT, Université Paul-Sabatier,
 \textsuperscript{3}Google Research,\\
 \textsuperscript{4}CerCo, CNRS, Université de Toulouse,
 \textsuperscript{5}ANITI, Université de Toulouse
\\
 \small{
   \textbf{Correspondence:} \href{mailto:antonin.poche@irt-saintexupery.com}{antonin.poche@irt-saintexupery.com}
 }
}

\begin{document}
\maketitle

% ======== %
% Abstract %
% ======== %
\begin{abstract}

    Concept-based explanations work by mapping complex model computations to human-understandable concepts. Evaluating such explanations is very difficult, as it includes not only the quality of the induced \textit{space of possible concepts} but also how effectively the chosen concepts are \textit{communicated} to users. Existing evaluation metrics often focus solely on the former, neglecting the latter.
    
    We introduce an evaluation framework for measuring concept explanations via \textit{automated simulatability}: a simulator's ability to predict the explained model's outputs based on the provided explanations. This approach accounts for both the concept space and its interpretation in an end-to-end evaluation. Human studies for simulatability are notoriously difficult to enact, particularly at the scale of a wide, comprehensive empirical evaluation (which is the subject of this work). We propose using large language models (LLMs) as simulators to approximate the evaluation and report various analyses to make such approximations reliable. Our method allows for scalable and consistent evaluation across various models and datasets. We report a comprehensive empirical evaluation using this framework and show that LLMs provide consistent rankings of explanation methods. Code available on \href{https://github.com/AntoninPoche/ConSim}{GitHub}.

\end{abstract}

\begin{figure}[t]
    \centering
    \includegraphics[width=0.48\textwidth]{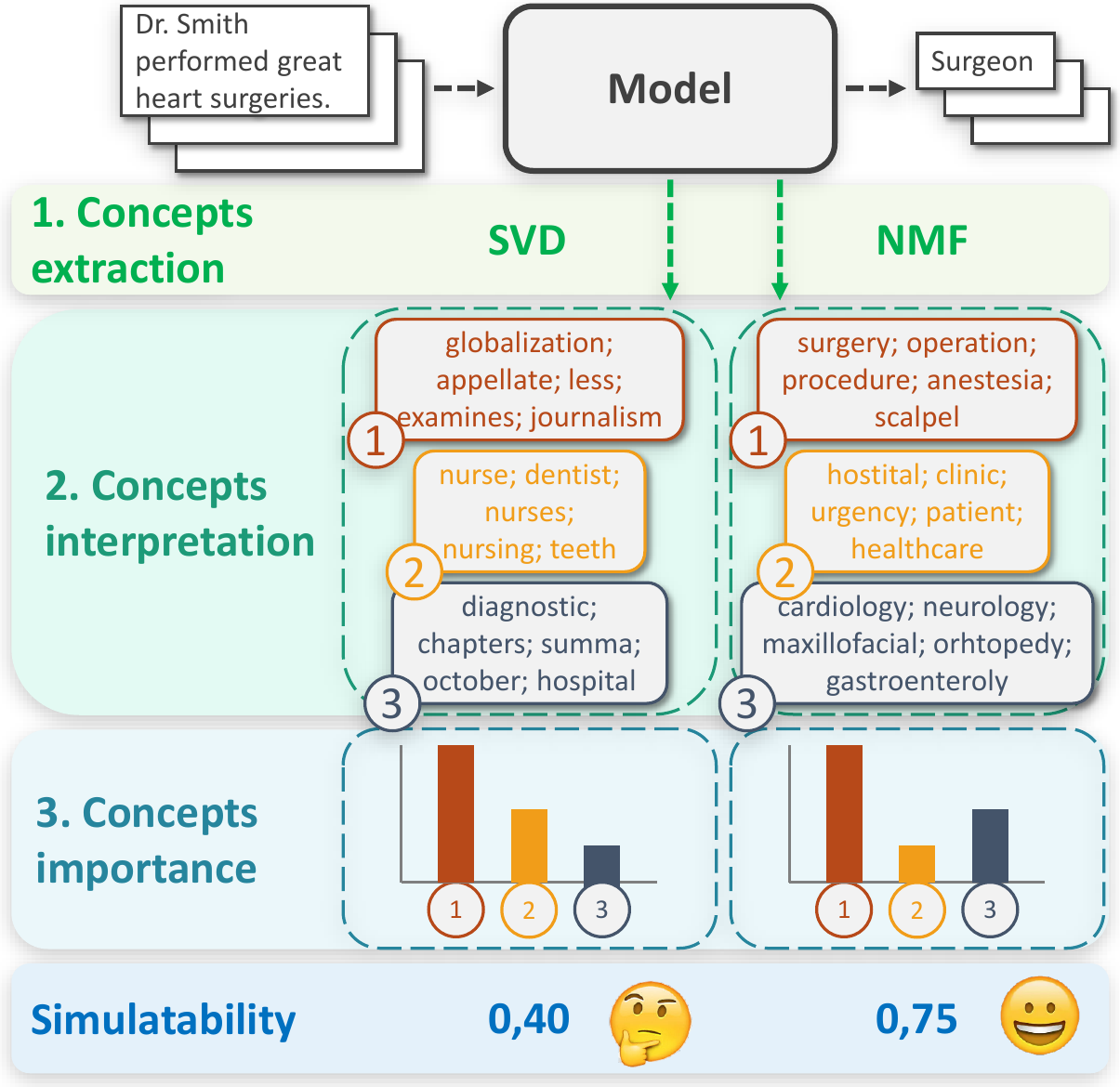}
    \caption{\textit{How can we choose concept extraction (1) and interpretation methods (2) to make them more useful to humans?} Concept-based XAI relies on identifying relevant, interpretable concepts in the model's latent space. Different techniques yield varying concepts and importance scores (3). The simulatability score (bottom) evaluates how effectively these explanations help users understand model predictions.
    \vspace{-5mm}}
    \label{fig:interpretation}
\end{figure}

% ============ %
% Introduction %
% ============ %
\section{Introduction}
    % guidelines questions from Alon:
        % What is the question we are answering?
        % Why is this question worth asking?
        % Why is answering this question challenging and/or unsolved until now?
        % What is the answer?
        % What did we do to answer the question and why is our methodology sound?

    % XAI
    The need for transparent and interpretable models has remained a principal need in NLP, leading to the emergence of Explainable AI (XAI) as a means of fostering trust and understanding in these systems. Among the various XAI approaches, concept-based explanations stand out for their ability to bridge the gap between complex model computations and human-understandable concepts. Unlike feature attribution methods that focus on individual input features, concept-based explanations group features into higher-level abstractions or "concepts" more aligned with human cognition \cite{deveaud2014accurate, kim2018interpretability,ghorbani2019towards,fel2023craft}, facilitating better interpretation of the model's internal reasoning. 

    % Concept-based Evaluation
    However, evaluating such methods remains challenging. Many evaluation metrics lack a basis in human interpretation and measure either faithfulness \cite{jacovi2020towards} or complexity. For example, as shown in Fig.~\ref{fig:interpretation}, SVD achieves higher scores across multiple metrics but produces less useful concepts for understanding model predictions. Additionally, current metrics prioritize concept-space evaluation while neglecting concept interpretation.
    Following previous work~\cite{fel2023holistic}, we argue that concept-based explanation frameworks have three main components: i) constructing the concept space, ii) evaluating concept importance, and iii) interpreting concepts.

    % Simulatability
    We propose using simulatability~\cite{hase2020evaluating,colin2022cannot} as a reliable method of enacting a comprehensive evaluation. Simulatability assesses the ability of a meta-predictor $\Psi$ (simulator) to understand predictions of a model $f$ by measuring the capacity of $\Psi$ to simulate the predictions of $f$ empirically. This end-to-end approach evaluates the usefulness of explanations \cite{colin2022cannot}.

    % Concept-based Evaluation (preprint version)
    % However, evaluating such methods remains a challenge. Evaluation metrics often lack grounding in human interpretation (e.g., see Fig.~\ref{fig:interpretation}--while SVD has a much higher score with many metrics, it leads to concepts that are much less useful for understanding the model's predictions). Current metrics are proxies for either faithfulness or plausibility \cite{jacovi2020towards}, and the trade-off between the two is rarely explored in this setting. Furthermore, existing metrics focus on concept-space evaluation and overlook the interpretation of concept dimensions.
    % Following previous work~\cite{fel2023holistic}, we argue that concept-based explanation frameworks have three main components: constructing the concept space, evaluating concept importance, and interpreting concepts.

    % Simulatability (preprint version)
    % We propose using simulatability~\cite{hase2020evaluating,colin2022cannot} as a reliable method of enacting a comprehensive evaluation. Simulatability assesses the ability of a meta-predictor $\Psi$ (simulator) to understand predictions of a model $f$ by measuring the capacity of $\Psi$ to simulate the predictions of $f$ empirically. This approach evaluates both faithfulness and plausibility.

    % Simulatability steps
    A simulatability experiment consists of three phases: i) $\Psi$ is introduced to the task during the Initial Phase (IP); ii) learns the model’s behavior in the Learning Phase (LP); and iii) attempts to simulate $f$’s predictions during the Evaluation Phase (EP). We adapted simulatability to concept-based explanations, optionally introducing model-wise explanations at IP and sample-wise explanations at LP. However, explanations should never be provided at EP so that the labels are not leaked.

    % User-LLM
    Simulatability is often evaluated through user studies. However, the number of participants necessary for an extensive method benchmark makes such studies prohibitively costly \cite{poursabzi2021manipulating,de2024evaluating} and notoriously sensitive to superficial confounders~\cite{10.1145/3531146.3533127}. In this paper, we explore using large language models (LLMs) as meta-predictors, referred to as user-LLMs. Previous work~\cite{de2024evaluating} has shown their potential
    %of such meta-predictors,
    with results exhibiting high correlations with human performance.

    % Our experiments and results
    We experimented with various datasets, models, user-LLMs, and methods. As simulatability scores are only comparable for equivalent settings, we aggregated these scores using Copeland's ranked-choice voting method \cite{copeland1951reasonable,szpiro2010numbers}. This gave us comparable method rankings regardless of the experimental setup. Furthermore, most of the differences between the pairwise methods were statistically significant. We have tested six different methods across various datasets and meta-predictors. Non-negative Matrix Factorization~\cite[NMF;][]{lee1999learning} was overall the best-performing method.

    % Contributions
    \paragraph{Contributions:}
    
    \paragraph{A generalizing formalization} of concept-based explanations components: (1) concept space, (2) concept importance, and (3) concept interpretation.
    
    \paragraph{An evaluation framework} using simulatability to assess the interpretability of concept-based explanation methods.
    
    \paragraph{User-LLMs for simulatability:} A demonstration of user-LLMs as effective meta-predictors in a simulatability framework.
    
    \paragraph{A comprehensive empirical analysis} across multiple use-cases, with statistical significance.

% ============ %
% Related work %
% ============ %
\section{Concept Explanations: Background}
    % ---
    % XAI
    The field of explainable artificial intelligence (XAI) for classification tasks has witnessed significant growth, driven by the widespread adoption of deep learning techniques. Among the various approaches, attribution methods \cite{zeiler2014visualizing,ribeiro2016should,shrikumar2017learning,lundberg2017unified} have traditionally dominated the literature, offering insights by highlighting the contributions of input features to model predictions. However, concept-based methods \cite{kim2018interpretability,ghorbani2019towards,koh2020concept,yeh2020completeness,zarlenga2022concept,jourdan2023cockatiel} have recently gained increasing attention, providing a complementary perspective by focusing on high-level, human-interpretable concepts to explain model behavior.

    % --------
    % Concepts
    \paragraph{Supervised vs. Unsupervised.} Within concept-based explainability methods, two main categories can be identified. The first relies on supervised concepts constructed using labeled concept datasets. This category includes methods such as CAV (Concept Activation Vector) \cite{kim2018interpretability} for post-hoc approaches and CBM (Concept Bottleneck Model) \cite{koh2020concept} for by-design frameworks. However, finding labeled concepts is inherently difficult, and creating datasets to represent concepts often introduces substantial human bias~\cite{ramaswamy2023overlooked}.
    
    By contrast, unsupervised concept-based methods extract concepts directly from the model's latent space. Neurons are not interpretable in themselves \cite{elhage2022toymodelsof,colin2024local,dreyer2024pure}. Hence, the most widely used approach treats concepts as a linear combination of neurons -- directions in the latent space \cite{kim2018interpretability,yeh2020completeness,zhang2021invertible,cunningham2023sparse,fel2023craft,jourdan2023cockatiel,zhao2024explaining}. Recent advances in mechanistic interpretability have focused on Sparse Auto-Encoders \cite[SAEs;][]{ng2011sparse,makhzani2013k,domingos2015master} to find these concepts \cite{bricken2023monosemanticity,rajamanoharan2024improving,rajamanoharan2024jumping,templeton2024scaling,gao2024scaling,Lieberum2024gemma,fel2024understanding}.
    % Intervention
    %Moreover, obtaining interpretable spaces might not be the goal, but a way. Some methods propose to intervene on concepts to modify models to remove unwanted biases \cite{jourdan2023taco,arditi2024refusal,gao2024scaling,marks2024sparse,hong2024intrinsic,bhalla2024towards}.

    % ---------------
    % Classic metrics
    \paragraph{Evaluations.} Post-hoc, unsupervised concept-based explanations evaluation typically focuses on two main properties: faithfulness \cite{jacovi2020towards} and complexity. Faithfulness-oriented metrics -- such as completeness \cite{yeh2020completeness}, fidelity \cite{zhang2021invertible}, relative $\ell_2$ \cite{fel2023holistic}, FID and OOD \cite{fel2023holistic}, and MAE \cite{bricken2023monosemanticity} -- measure how well the identified concepts preserve the information from the model’s original embeddings. In addition, complexity is often inferred from proxies such as sparsity \cite{fel2023holistic,bricken2023monosemanticity} and conciseness \cite{vielhaben2023multi}.
    
    % ---------
    % Benchmark
    Many evaluation frameworks rely on labeled concepts (e.g., CEBaB \cite{abraham2022cebab}), which are often challenging to define, validate, and align with a model’s internal representations. Although some studies have performed human evaluations \cite{zhang2021invertible,barua2024concept}, to the best of our knowledge, no previous work has applied simulatability to concept-based explanations.

    % --------------
    % Simulatability
    \paragraph{Simulatability.} Simulatability refers to the extent to which ``an user can correctly and efficiently predict the method’s results'' \cite{kim2016examples,hase2020evaluating,colin2022cannot}. It assesses how useful and understandable an explanation is to a user. Recent studies suggest that large language models (LLMs) can approximate human judgments at scale \cite{de2024evaluating,nguyen2024interpretable}. Some works use LLMs as meta-predictors to evaluate simulatability automatically. For instance, \citet{mills2023almanacs} examine a wide range of explanations but do not cover concept-based explanations. Additionally, \citet{chen2024models} introduce counterfactual simulatability to assess generalization around a decision, while our framework evaluates generalization across all samples. Finally, \citet{chan2022frame} argue that the aspect of simulatability being measured depends on the chosen meta-predictor.

\section{A Theoretical Framework for Post-hoc Unsupervised Concept-XAI}

\begin{figure}[tp]
        \centering
        \includegraphics[width=0.45\textwidth]{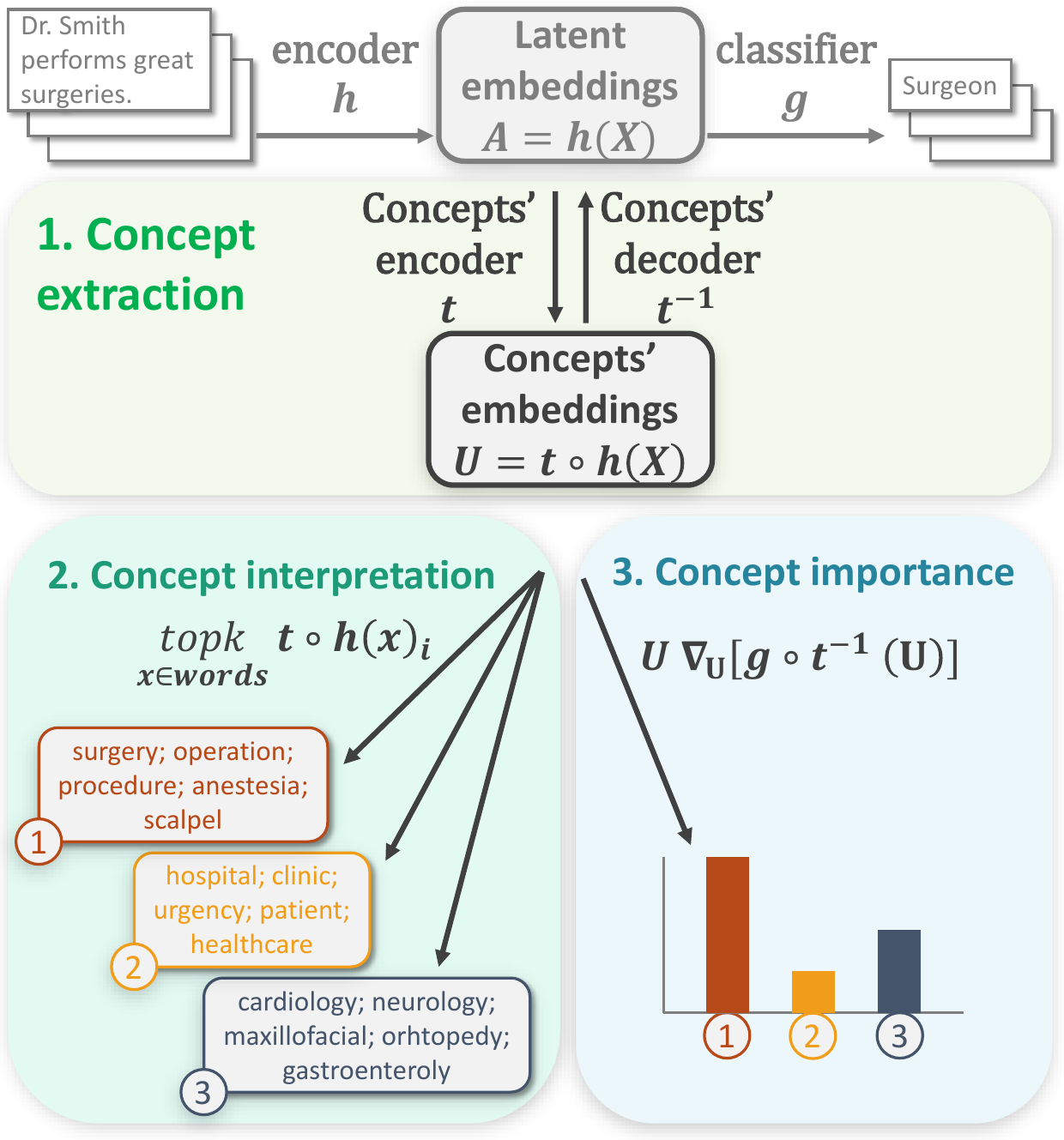}
        \caption{\textit{A generalizing formalization of Concept-based explanations.} For a model $f = g \circ h$, concepts can be extracted from its activations $A = h(X)$ using the \textit{concept encoder} $t$, and can be decoded using the \textit{concept decoder} $t^{-1}$. The explanation can interpreted by keeping the most relevant words for each concept. Finally, an importance score can be attributed to each concept to understand their role in the model's rationale.
        \vspace{-8mm}
        }
        \label{fig:concepts_diagram}
    \end{figure}

    Consider classification models $f: \mathcal{X} \longrightarrow \mathcal{Y}$ with input space $\mathcal{X}$ and output space $\mathcal{Y}$. The model is decomposed into: $f = g \circ h: \mathcal{X} \xrightarrow[]{h} \mathcal{H} \xrightarrow[]{g} \mathcal{Y}$ with $\mathcal{H} \subseteq \mathbb{R}^p$ the embedding space. In our experiments, we divide the model at the penultimate layer, in DistilBERT~\cite{sanh2019distilbert}, $h$ outputs would be the token [CLS]. Concept-based explanations have three main components described in the following three subsections and illustrated in Fig.~\ref{fig:concepts_diagram}.

    % ---------------------
    % Define concepts space
    \subsection{Concepts Space}
    
        The first step of post-hoc unsupervised concept-based explainability is to define the concept space $\mathcal{C} \subseteq \mathbb{R}^k$ through concept extraction methods. \textit{Concept extraction methods} allow the construction of a projection $t: \mathcal{H} \longrightarrow \mathcal{C}$ and its bijection (or approximation) $t^{-1}: \mathcal{C} \longrightarrow \mathcal{H}$ (appendix \ref{appendix:methods} defines how are obtained such concept projection). The input-to-concept part is noted $f_{ic}: \mathcal{X} \xrightarrow[]{h} \mathcal{H} \xrightarrow[]{t} \mathcal{C}$ and the concept-to-output part $f_{co}: \mathcal{C} \xrightarrow[]{t^{-1}} \mathcal{H} \xrightarrow[]{g} \mathcal{Y}$. Finally, we can construct $f_c = f_{co} \circ f_{ic}: \mathcal{X} \xrightarrow[]{f_{ic}} \mathcal{C} \xrightarrow[]{f_{co}} \mathcal{Y}$, an unsupervised CBM (Concept-Bottleneck Model) \cite{koh2020concept}.
        Concepts extraction is done class by class in many concept-based explainability methods for classification. However, we treat all classes simultaneously to obtain a common concept space, as in \citet{jourdan2023taco}.

    % --------------------
    % Communicate concepts
    \subsection{Concepts Interpretability} \label{interpretability}

        The second part of post-hoc unsupervised concept-based explainability is to interpret concepts. Concepts are directions in the latent space and are not interpretable as is. How to represent a concept is still an open question. It is possible to represent concepts as word clouds \cite{dalvi2022discovering}, give examples that activate the concepts and highlight important words \cite{jourdan2023cockatiel}, or label the given concepts by either: asking human annotators \cite{dalvi2022discovering}; finding the most aligned label in a concept bank \cite{sajjad2022analyzing}; or asking an LLM to label the concept based on maximally activating examples \cite{bricken2023monosemanticity,templeton2024scaling}. The last solution has been the most popular in the mechanistic interpretability literature. However, its computational cost is high for interpreting a single concept. In this paper, we explored two different interpretability methods:
        
        \paragraph{Concept Maximally Activating Words (CMAW)} selects the five words that most strongly activate a concept and, if negative activations exist, also the five least activating words. These words are selected from words frequent enough in the dataset. With regards to concept dimension $i$, CMAW can be computed as follows:
        \begin{equation}
            CMAW(cpt_i) = \topk_{x \in words} f_{ic}(x)_i
            \label{eq:cmaw}
        \end{equation}

        \paragraph{o1 Concept Alignment (o1CA)}. For o1CA, we prompt GPT o1 \cite{openai2024introducing} for potential concept labels and corresponding representative sentences, then align discovered concepts to these labels by choosing the label with the highest mean activation on the corresponding sentence. Thus, for our concept dimension $i$, with $X_j$ the sentences corresponding to o1 concept $j$, we have:
        \begin{equation}
            o1CA(cpt_i) = \max_{j \in o1\_cpt} \mean_{x \in X_j} f_{ic}(x)_i
            \label{eq:o1ca}
        \end{equation}

    % -------------------
    % Concepts importance
    \subsection{Concepts Importance}
    
        Concept attribution methods $\varphi: \mathcal{C}  \longrightarrow \mathbb{R}^k$ provide the importance of each concept for a given prediction based on the concepts. \citet{fel2023holistic} show (theorem 3.2) that when the model is divided at the penultimate layer, certain attribution methods (e.g., Gradient Input \cite{shrikumar2017learning}) are optimal. We, therefore, choose Gradient Input for its simplicity and efficiency. \textit{Local concepts' importance} $\varphi$ can be defined for a given sample $x \in \mathcal{X}$, with concepts representation $u = f_{ic}(x) \in \mathcal{C}$, by Eq.~\ref{eq:local_explanation}. Through this, with $X \in \mathcal{X}^n$ the train set samples and $U = f_{ic}(x) \in \mathcal{C}^n$ their concepts representations, we can define \textit{global concepts importance} $\varPhi$ with regard to class $c$ through Eq.~\ref{eq:global_explanation}.
        \begin{equation}
            \varphi_{f_{co}}(u) =\ u \nabla_u f_{co}(u)
            \label{eq:local_explanation}
        \end{equation}
        \begin{equation}
            \varPhi_{f_{co},c} = \mean_{u \in U | f_{co}(u) = c} \varphi_{f_{co}}(u)
            \label{eq:global_explanation}
        \end{equation}

% ======================== %
% Our Evaluation Framework %
% ======================== %
\section{Our Evaluation Framework} \label{benchmark}

    \begin{figure*}[tp]
        \centering
        \includegraphics[width=\textwidth]{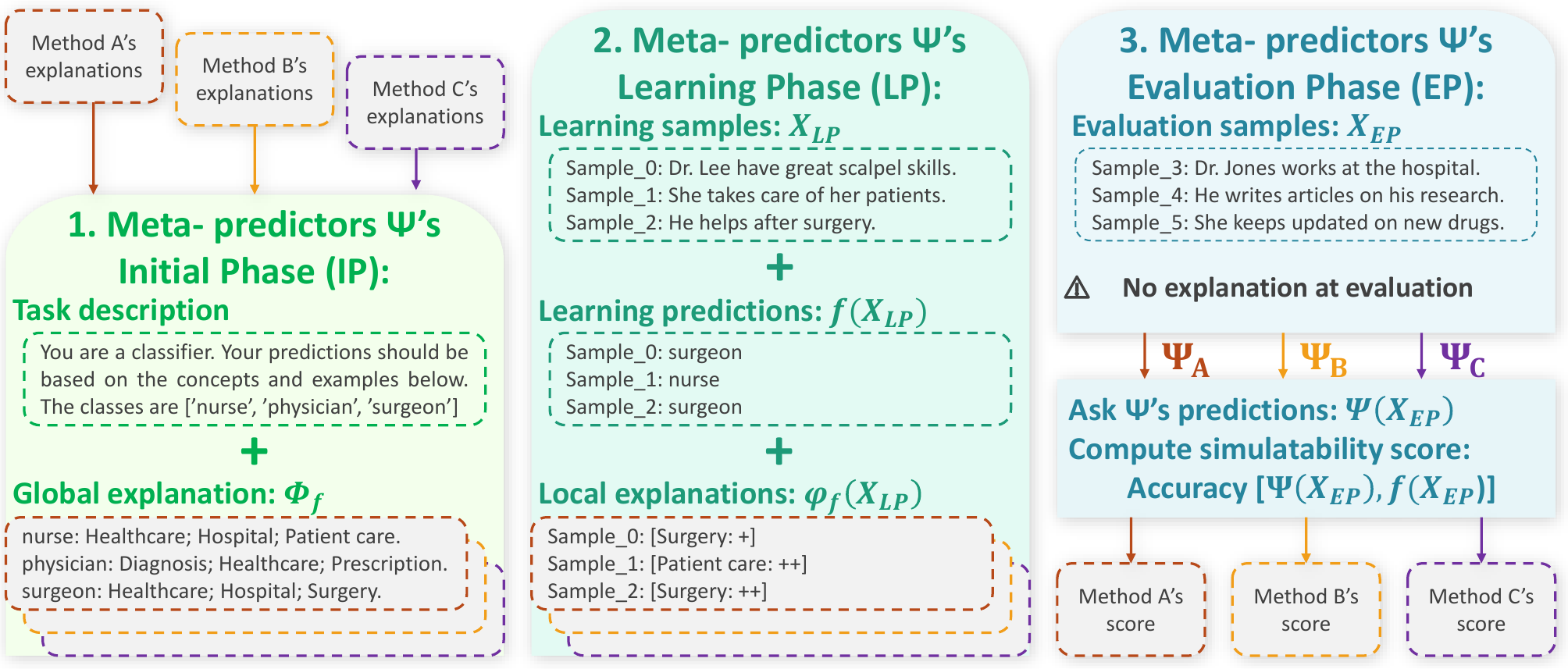}
        \caption{\textit{Overview of our simulatability framework.} For a given \textit{meta-predictor} $\Psi$ (User-LLM or human), our simulatability framework is composed of three distinct stages: \textit{(i)} an \textit{initial phase} (\textit{IP}) where the task is carefully described to $\Psi$ and the global explanation is shown to it; \textit{(ii)} a \textit{learning phase} (\textit{LP}) where some samples are shown to $\Psi$, along with the model $f$ predictions; \textit{(iii)} a final \textit{evaluation phase} (\textit{EP}) where a different set of samples is input to $\Psi$ without the corresponding predictions, and it is asked to predict what the model $f$ would have predicted. With this information, the simulatability score can be computed as the accuracy in guessing the model's outputs.
        \vspace{-3mm}
        }
        \label{fig:framework_diagram}
    \end{figure*}

    % --------------
    % Simulatability
    \subsection{Simulatability} \label{simulatability}

        Simulatability aims to quantify how well a meta-predictor $\Psi$ (also called simulator) can replicate the predictions of an AI model $f$ \cite{kim2016examples,hase2020evaluating,colin2022cannot}. The meta-predictor is usually a human, but in our experiments, we use an LLM as a meta-predictor. The meta-predictor is given samples and tasked to predict what the AI model would have predicted.
        
        A simulatability experiment consists of three phases. These parts are illustrated in Fig.~\ref{fig:framework_diagram} through reduced examples of prompt parts:
        
        \paragraph{Initial Phase (IP):} The meta-predictor receives a description of the task and possibly some global explanations of the model. Global explanations consist of global concepts' importance $\varPhi$ as defined by Eq.~\ref{eq:global_explanation} and the important concepts' interpretation.
        
        \paragraph{Learning Phase (LP):} The meta-predictor is shown examples with the model’s predictions and, optionally, local explanations. The explanations are concepts' importance $\varphi$ as defined by Eq.~\ref{eq:local_explanation}.
        
        \paragraph{Evaluation Phase (EP):} The meta-predictor must predict the model’s outputs on new samples without access to these predictions. No explanation is given at this phase as it would leak the label.

\vspace{0.2cm}
        \noindent In summary, $\Psi$ is introduced to the task during IP, learns the model’s behavior in LP, and attempts to simulate $f$’s predictions in EP. Since $\Psi$'s performance may depend on the experimental settings $s$ and the chosen concept extraction method $m$, we denote it as $\Psi_{s,m}$. By assessing how accurately $\Psi$ replicates $f$’s outputs, this approach mitigates issues like confirmation bias and prediction leakage \cite{colin2022cannot}. We measure simulatability as the accuracy of the meta-predictor’s guesses on EP samples $X_{EP}$:
        \begin{equation}
            acc_{\Psi,s,m} = \expectation\limits_{x \in X_{EP}} \mathds{1}\left\{\Psi_{s, m}(x_i) = f(x_i)\right\}
            \label{eq:acc}
        \end{equation}
        % \begin{align}
        %     acc_{\Psi,s,m} = \frac{1}{N} \sum_{i = 0}^N &\mathds{1}\left\{\Psi_{s, m}(x_i) = f(x_i)\right\} \\ 
        %     X_{EP} &= \{x_i\}_0^N
        % \end{align}

        In each setting, samples for LP and EP were selected to represent the dataset and better differentiate methods explaining performance. Each setting had different seeds for more statistically significant results. Details are described in appendix \ref{appendix: prompt_samples}.

    % ----------------------
    % User-LLM and Prompting
    \subsection{User-LLM and Prompting}

        We refer to LLMs replacing users in user studies as user-LLMs \cite{de2024evaluating}. User-LLMs do not replace studies with real humans, but they allow experiments at a much larger scale to provide an approximation and motivation for future investment in human user studies. Furthermore, it was shown the conclusions of studies through the lens of user-LLMs tend to correlate with human studies \cite{de2024evaluating}.
        
        In our case, we leverage GPT-4o-mini \cite[\S\ref{GPT-4o-mini_xp}]{openai2024openai} and Gemini-1.5 Flash and Pro \cite[\S\ref{user_llms_xp}]{team2024Gemini}. The Gemini experiments cover a representative subset of the full experiment scope.

        \paragraph{Selecting the concepts.} Some of the induced concept spaces had 500 concepts. Showing them all would complicate the prompt unnecessarily. Therefore, for global and local explanations, we only show concepts with normalized global importance $\hat{\varPhi}_{c,cpt}$ in absolute value above a threshold of $0.05$ for at least one class. For a given class $c$ and concept $cpt$, its normalized global importance is defined in appendix \ref{appendix: normalize_importance}, Eq.~\ref{appendix:eq:normalized_global_importance}. Similarly, for the remaining concepts, local explanations only include concepts with importance values above $0.05$ for the given sample. Normalized local importance is defined in appendix \ref{appendix: normalize_importance}, Eq.\ref{appendix:eq:normalized_local_importance}. In prompts, concepts' importance is encoded into four buckets for simplicity; details in appendix \ref{appendix:communicate_importance}. An example of a prompt can be found in the appendix~\ref{appendix:prompt_example}.

    % -------
    % Ranking
    \subsection{Ranking}

        Different settings are constructed by fixing the dataset, model, seed, concepts' extraction method, concepts' interpretation method, prompt type, and user-LLM. However, the simulatability scores $acc_{\Psi,s,m}$ (Eq.~\ref{eq:acc}) between the two methods can only be compared when the setting $s$ is the same. Therefore, to rank methods, we make the parallel with ranked-choice voting systems. We consider the simulatability score from a setting as a vote with order between methods and aggregate these votes using Copeland's method \cite{copeland1951reasonable,szpiro2010numbers} with the "$0/1/2$" rule, a kind of Condorcet method \cite{pomerol2012multicriterion}. Afterward, with $S$ the settings, $i$, and $j$ methods, we construct the pairwise comparison matrix $P$ through Eq.~\ref{eq:pairwise_percentage}. Note that here, "methods" are either concept extraction methods, concept interpretation methods, or concept importance methods.
        \begin{equation}
            P_{i,j} = \sum_{s \in S} \begin{cases}
                0 & \text{if } acc_{\Psi,s,i} < acc_{\Psi,s,j} \\
                1 & \text{if } acc_{\Psi,s,i} = acc_{\Psi,s,j} \\
                2 & \text{if } acc_{\Psi,s,i} > acc_{\Psi,s,j}
            \end{cases}
            \label{eq:pairwise_percentage}
        \end{equation}

        Each value is then normalized to obtain a value between 0 and 100 comparable to a percentage of wins. Finally, the ranking of a method $i$ is constructed from the number of times method $i$ is preferred over method $j$, with $M$ the list of concepts explanation methods:
        \begin{equation}
            rank_P(i) = |M| + 1 - \sum_{j \in M} \mathds{1}\left\{P_{i,j} \geq 50\right\}
            \label{eq:ranking}
        \end{equation}
        
        Furthermore, another pairwise comparison matrix was computed to determine if the pairwise differences were statistically different than $0$. We used a student's test \cite{student1908probable} with a p-value threshold of $0.05$. To do so, the mean differences between accuracies were computed with:
        \begin{equation}
            Diff_{i,j} = \mean_{s \in S} [acc_{\Psi,s,i} - acc_{\Psi,s,j}]
            \label{eq:pairwise_difference}
        \end{equation}

% ================================ %
% Extended GPT-4o-mini Experiments %
% ================================ %
\section{Ranking Methods with GPT-4o-mini}

    The first experiment was conducted with GPT-4o-mini~\cite{openai2024openai} as meta-predictor $\Psi$. Comparison of the ranking between several user-LLMs are described in Sec.~\ref{user_llms}.

    % -----------------------------------
    % GPT-4o-mini Experiments Description
    \subsection{GPT-4o-mini Experiments Description}  \label{GPT-4o-mini_xp}

        Experiments with GPT-4o-mini were conducted with an extended set of settings compared to the latter comparison. We use 4 datasets, 5 models, 7 seeds, 5 concept extraction methods, 2 concept interpretation methods, 6 prompt types for explanations, 4 other prompt types for baselines, and, for some settings, 7 different numbers of concepts. There are also several baseline prompts. Resulting in $24,560$ different experiment settings reported and used for GPT-4o-mini. Prompt mean size was about $2,000$ tokens; hence, through the OpenAI API, this cost around $7\$$. The different settings variables are listed below:
        
        % (4*5*5*5*2*6 = 6000) + (4*5*2*5*2*6*7 = 16800) + (4*5*7*4 = 560) + (4*5*5*1*2*6*1 = 1200)
    
        \paragraph{Datasets.} We consider four classification datasets:  
            (i) A reduced version of BIOS \cite{de2019bias}, limited to the 10 most frequent classes;  
            (ii) IMDB \cite{maas2011learning};  
            (iii) Rotten Tomatoes \cite{pang2005seeing};  
            (iv) The "emotion" subset of the Tweet Eval dataset \cite{barbieri2020tweeteval}.
        For concept extraction, we often augment the original datasets by including split samples (partial sentences) derived from the initial samples. See extended details in Appendix \ref{appendix:datasets}.
        
        \paragraph{Models.} We evaluate three model architectures:  
            an encoder model DistilBERT \cite{sanh2019distilbert},  
            an encoder-decoder model T5 \cite{raffel2020exploring}, and  
            a decoder model Llama3-8B \cite{dubey2024llama}.
        DistilBERT and T5 were fine-tuned for the classification tasks, while Llama3-8B used prompting. Details of model fine-tuning and adaptation are in appendix \ref{appendix:models}. DistilBERT and T5 were fine-tuned with positive embeddings to enable NMF-based concept extraction. These modified models are denoted with $+$ in Tab.~\ref{tab: GPT-4o-mini_res} and Tab.~\ref{tab:user_llm_comparison}, resulting in five distinct models in total.
        
        \paragraph{Concept extraction methods.} We employed five concept extraction methods, each representing a form of dictionary learning as generalized in \cite{fel2023holistic} and considered not making any projection as a baseline:  
            (i) Independent Component Analysis (ICA) \cite{ans1985architectures,hyvarinen2000independent};
            (ii) Non-negative Matrix Factorization (NMF) \cite{lee1999learning,sra2005generalized};
            (iii) Principal Component Analysis (PCA) \cite{pearson1901liii,hotelling1992relations};
            (iv) Sparse Auto-Encoder (SAE) \cite{ng2011sparse,makhzani2013k,domingos2015master};  
            (v) Singular Value Decomposition (SVD) \cite{eckart1936approximation}, and
            (vi) the \textit{NoProjection} baseline \cite{geva2022transformer}.
        See appendix \ref{appendix:methods} for details on their implementation and the corresponding notation.

        The \textit{NoProjection} baseline method is evaluated with explanation, but the concept space corresponds to the latent embedding space $\mathcal{C} = \mathcal{H}$. While for the \textit{NoExplanation}, no explanations are shown to the meta-predictor $\phi = \Phi = \varnothing$.
        
        \paragraph{Concept interpretation methods.} Experiments use the two concept interpretability methods introduced in Section~\ref{interpretability}, namely CSAW and o1CA.
        
        \begin{table}[tp]
            \centering
            \ra{1.2}
            \small
            \begin{tabular}{@{}llccccc@{}}
                \toprule
                \multicolumn{2}{l}{\textbf{Simulatability Phase}}         & \textbf{NE1}  & \textbf{E1}  & \textbf{NE2}  & \textbf{E2}  & \textbf{E3}  \\
                \midrule
                \multirow{2}{*}{IP} & Task desc.                & \cm & \cm & \cm & \cm & \cm \\
                                    & Global expl.              &     & \cm &     & \cm & \cm \\ \midrule
                \multirow{2}{*}{LP} & $X_{LP}$, $f(X_{LP})$     &     &     & \cm & \cm & \cm \\
                                    & Local expl.               &     &     &     &     & \cm \\
                \bottomrule
            \end{tabular}
            \caption{Different simulatability prompting experiments characterized by the elements inside. E1: No Learning; E2: Only global explanation; and E3: Both local and global explanations. NE1 and NE2 are the corresponding \textit{NoExplanation} baselines. IP: Initial Phase; LP; Learning Phase. Details in Sec.~\ref{simulatability} and Fig.~\ref{fig:framework_diagram}.
            % caption preprint version
            % \caption{Elements present in the simulatability prompt depending on the experiment (E1, E2, or E3). or the baseline (L1 or L2). Details in Sec.~\ref{simulatability} and Fig.~\ref{fig:framework_diagram}.
            \vspace{-3mm}
            }
            \label{tab:prompt_types}
        \end{table}

        \begin{table*}[t]
            \centering
            \scriptsize
            \ra{1.2}
            \begin{tabular}{@{}lcccccccccccc@{}}
                \toprule
                \multicolumn{2}{c}{\textbf{Experiment setting subset}}  &  \multicolumn{7}{c}{\textbf{Concept extraction}}                    && \multicolumn{3}{c}{\textbf{Concept interpretation}}\\
                \cmidrule(lr){3-9} \cmidrule(lr){11-13}
                                                                        &&  NMF  &  SAE   &  ICA   & NoProjection & NoExplanation & PCA & SVD &&  CMAW  &  o1CA  & NoExplanation \\
                \midrule
                                    
                \multirow{4}{*}{Datasets}       & BIOS10                & \tb{1} & \ul{2} &     3  &     4        &     7         &  5  &  6  && \tb{1} & \ul{2} &     3  \\
                                                & IMDB                  & \tb{1} &     5  &     3  &     3        & \ul{2}        &  7  &  6  && \tb{1} &     3  & \ul{2} \\
                                                & rotten tomatoes       & \tb{1} & \ul{2} &     4  &     4        &     3         &  7  &  6  && \tb{1} & \ul{2} &     3  \\
                                                & tweet eval            & \tb{1} &     4  &     3  & \ul{2}       &     5         &  6  &  7  && \tb{1} & \ul{2} &     3  \\
                \midrule
                \multirow{5}{*}{Models}         & DistilBERT            &    N/A & \tb{1} & \tb{1} &     4        &     3         &  5  &  6  && \tb{1} & \ul{2} &     3  \\
                                                & DistilBERT+           & \ul{2} &     3  &     2  & \tb{1}       &     6         &  5  &  7  && \tb{1} & \ul{2} &     3  \\
                                                & Llama-3-8B            &    N/A & \ul{2} & \tb{1} &     3        &     6         &  5  &  6  && \tb{1} & \ul{2} &     3  \\
                                                & T5                    &    N/A & \tb{1} & \ul{2} &     4        &     3         &  5  &  6  && \ul{2} & \tb{1} &     3  \\
                                                & T5+                   & \tb{1} &     3  &     4  & \ul{2}       &     5         &  6  &  7  && \tb{1} & \ul{2} &     3  \\
                \midrule
                All settings                    &                       & \tb{1} & \ul{2} & \ul{2} &     4        &     5         &  6  &  7  && \tb{1} & \ul{2} &     3  \\ 
                \bottomrule
            \end{tabular}
            \caption{\textit{Methods ranking with GPT-4o-mini.} Comparison of concept extraction methods and concept interpretation methods rankings across different sets of settings. In a setting (a line), we fix either one of the datasets or models. The last line shows the ranking for all settings of the extended GPT-4o-mini experiments. \textit{NoProjection}, PCA, and SVD are removed from the comparison of concept interpretation methods.
            \vspace{-3mm}
            }
            \label{tab: GPT-4o-mini_res}
        \end{table*}
        
        \paragraph{Prompt types.} \label{prompt_types} We explored several prompt configurations to answer questions, such as whether a learning phase (LP) improves user-LLM performance and whether local explanations are beneficial. Tab.~\ref{tab:prompt_types} details these prompt settings. The \textit{NoExplanation} prompts (NE1 and NE1) are baselines. E1 is compared to NE1, and settings with an LP is compared to NE2. All instructions are provided in the system prompt, and the evaluation phase (EP) is given in the user prompt. A complete prompt example is presented in appendix~\ref{appendix:prompt_example}.
        
        \paragraph{Anonymous prompt types.} While the user-LLMs ($\Psi$) can achieve performance levels close to those of the fine-tuned models on the initial task, our objective is for them to predict exactly what model $f$ would predict. To increase complexity, we introduced experiments where class labels are anonymized, \textit{e.g.} \texttt{"Surgeon"} becomes \texttt{"Class\_0"}. These experiments ensure that $\Psi$ must rely more on the provided concepts and explanations rather than directly recognizing class names.
        
        \paragraph{Number of concepts.} Finally, each concept extraction method includes a hyperparameter specifying the number of concepts. We tested $k\in\{3, 5, 10, 20, 50, 150, 500\}$ concepts. Some configurations timed out with a large $k$ (ICA and NMF). Instead of reporting results for every setting or always selecting the best outcome, we followed the validation procedure described at the end of Sec.~\ref{simulatability}, using two 40-sample sets to determine the optimal number of concepts for each dataset-method pair. The best number of concepts was often very high, which can be explained by the fact that we only showed the most important concepts.

        \paragraph{In summary,} a large variety of settings were explored to obtain statistically robust and generalizable results. This experiment has shown that NMF, SAE, and ICA are the most promising concept extraction methods. Furthermore, the concept interpretation method CMAW -- the simplest of the two -- is above o1CA in most cases.

        \begin{table*}[t]
            \centering
            \ra{1.2}
            \scriptsize
            \begin{tabular}{@{}lccccccccccc@{}}
                \toprule
                \multirow{2}{*}{\textbf{User-LLMs}}    &  \multicolumn{7}{c}{\textbf{Concept extraction}}  && \multicolumn{3}{c}{\textbf{Concept interpretation}}\\
                \cmidrule(lr){2-8} \cmidrule(lr){10-12}
                                    &  NMF   & NoProjection &  SAE   & ICA & PCA & NoExplanation & SVD &&  CMAW  &  o1CA  & NoExplanation \\
                \midrule
                GPT-4o-mini         & \tb{1} & \ul{2}       &     3  &  4  &  5  &     6         &  7  && \tb{1} & \ul{2} &     3    \\
                Gemini-1.5-flash    & \tb{1} &     3        & \ul{2} &  4  &  5  &     6         &  7  && \tb{1} & \ul{2} &     3    \\
                Gemini-1.5-pro      & \tb{1} & \ul{2}       &     3  &  4  &  5  &     7         &  6  && \tb{1} & \ul{2} &     3    \\
                \bottomrule
            \end{tabular}
            \caption{\textit{User-LLMs ranking comparison.} Comparison of concept extraction methods and concept interpretation methods rankings across different user-LLMs on the representative subset of experiments described in Sec.~\ref{user_llms_xp}.
            \vspace{-3mm}
            }
            \label{tab:user_llm_comparison}
        \end{table*}

    % -------------------
    % GPT-4o-mini Results
    \subsection{GPT-4o-mini Results}  \label{GPT-4o-mini_res}

        \begin{figure}[t]
            \centering
            \includegraphics[width=0.48\textwidth]{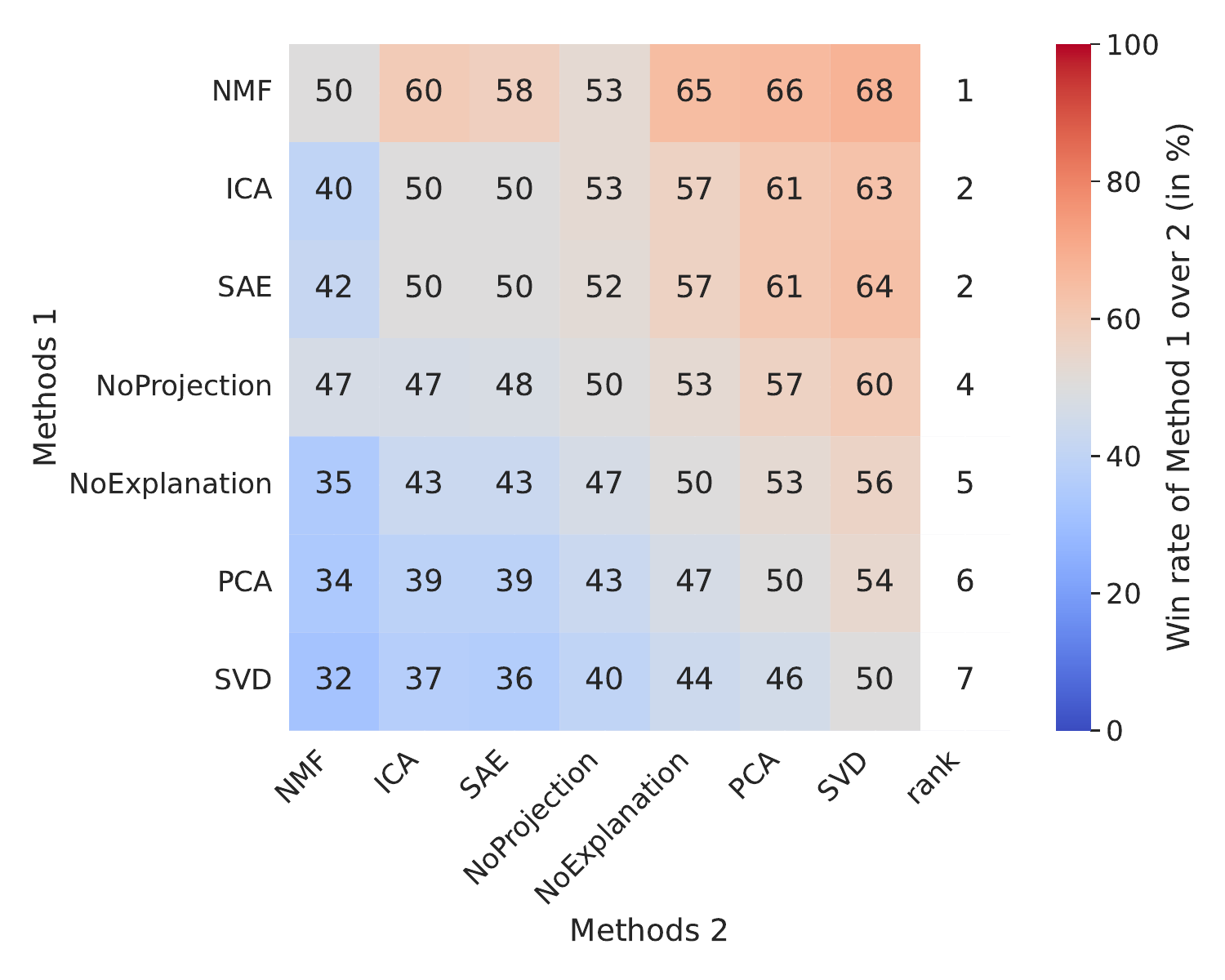}
            \caption{Pairwise comparison matrices on GPT-4o-mini experiments described in Sec.~\ref{GPT-4o-mini_xp}. Percentage of simulatability experiments where method 1 is over method 2. Ranking by number of pairwise victories.
            \vspace{-3mm}
            }
            \label{fig: GPT-4o-mini_percentage}
        \end{figure}

        GPT-4o-mini experiments can be analyzed from different angles: first, comparing concept extraction methods; second, comparing concept interpretation methods; and third, comparing the prompt types. In any case, results are primarily aggregated in pairwise comparison matrices Eq.~\ref{eq:pairwise_percentage} and Eq.~\ref{eq:pairwise_difference}, then the ranking is constructed following Eq.~\ref{eq:ranking}.

        \paragraph{Concept extraction methods.} Examples of pairwise comparison matrices defined in Eq.~\ref{eq:pairwise_percentage} and Eq.~\ref{eq:pairwise_difference} are respectively shown in Fig.~\ref{fig: GPT-4o-mini_percentage} and Fig.~\ref{appendix:fig: GPT-4o-mini_difference}. Fig.~\ref{fig: GPT-4o-mini_percentage} shows the percentage of wins between two methods and the final ranking of methods, putting the NMF above the others. Fig.~\ref{fig: GPT-4o-mini_percentage} shows that most differences are statistically significant with respect to a student's test \cite{student1908probable} with a p-value threshold of $0.05$.
        
        Tab.~\ref{tab: GPT-4o-mini_res} summarizes the ranking across settings with GPT-4o-mini as the user-LLM. The first line shows that overall, NMF ranks higher than SAE and ICA, which also rank higher than the baselines \textit{NoProjection} and \textit{NoExplanation}. Finally, the PCA and SVD rank, overall, below the baselines. Tab.~\ref{tab: GPT-4o-mini_res} also shows that the rankings are coherent with the general conclusion across subsets of settings with either one of the dataset, model, or concept interpretation methods fixed. Indeed, NMF, when applicable, is always ranked first except once, SAE and ICA occupy the top 3, but are sometimes overthrown by one of the baselines. Finally, PCA and SVD stay in the bottom three in any case. However, the baseline rank, thus the performance of methods overall, varies a lot with the dataset.

\begin{figure}[t]
            \centering
            \includegraphics[width=\linewidth]{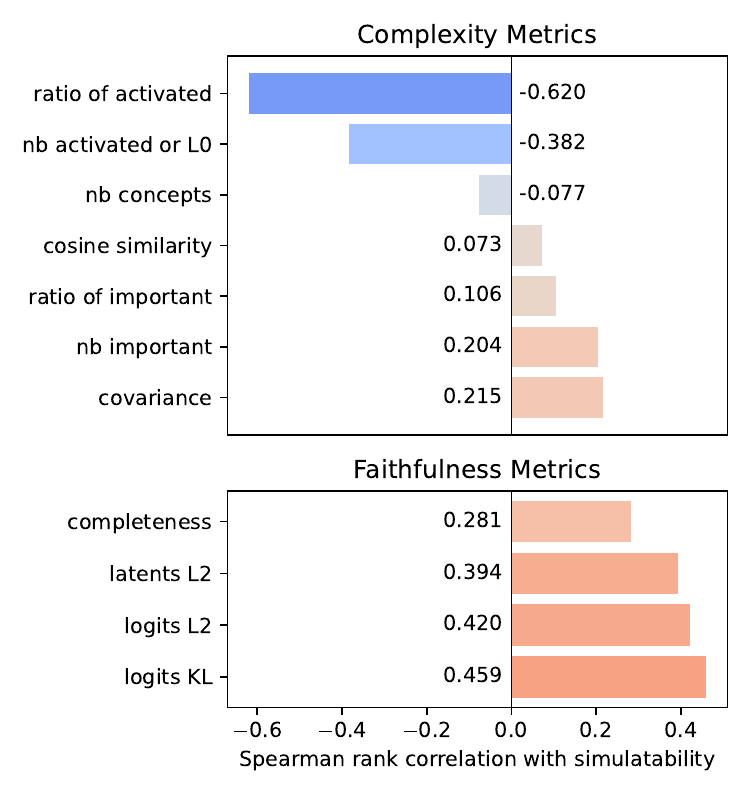}
            \caption{\textit{Spearman's rank correlation between simulatability and other concept-based metrics}. Other metrics are divided between faithfulness and complexity. It shows that higher fidelity and lower complexity tend to increase simulatability.
            \vspace{-5mm}}
            \label{fig: metrics correlations}
        \end{figure}

        \paragraph{Comparison with other metrics.} The Spearman's ranks correlation \cite{spearman1904proof} on concept extraction methods rankings between simulatability and eleven other metrics is illustrated in Fig.~\ref{fig: metrics correlations}. It shows that simulatability positively correlates with faithfulness metrics and negatively correlates with complexity metrics. This suggests that automated simulatability does not evaluate only faithfulness or complexity but what seems to be a trade-off between the two. Even if certain faithfulness and complexity metrics are themselves negatively correlated. Details on other metrics can be found in the appendix \ref{appendix: other metrics}.

        \paragraph{Concept interpretation method.} Concepts' interpretability methods were compared on the top 3 concept extraction methods (NMF, SAE, and ICA). Others were discarded due to their low results. Tab.~\ref{tab: GPT-4o-mini_res} shows that CMAW is better than o1CA and the \textit{NoExplanation} baseline. However, o1CA is behind the baseline for half of the datasets. Finally, Tab.~\ref{tab: GPT-4o-mini_res} shows consistent results across settings, with about half of them being statistically significant.

        \paragraph{Prompt types.} \label{GPT-4o-mini_prompt_types} The statistical differences between prompt types are illustrated in Fig.~\ref{appendix:fig: GPT-4o-mini_prompt_types}, where significant differences are in bold. It shows that no difference between settings with real class names can be made, suggesting that GPT-4o-mini shortcuts the task and ignores concepts. However, for setting anonymous classes representing more complex tasks, the GPT-4o-mini simulatability score obtains a clear gain with explanations. Finally, local explanations do not seem to help if global explanations are given.
        The same conclusions can be drawn from the distribution of concept extraction methods' simulatability score with respect to the prompt types in Fig.~\ref{appendix: GPT-4o-mini_distribution}.

% ================================ %
% User-LLMs Comparison Experiments %
% ================================ %
\section{User-LLMs Comparison Experiments} \label{user_llms}

    % --------------------------------------------
    % User-LLMs Comparison Experiments Description
    \subsection{Comparison Experiments Description} \label{user_llms_xp}
    
        For the second set of experiments, we compared the previous GPT-4o-mini results with more advanced user-LLMs, such as Gemini-1.5 (flash and Pro) \cite{team2024Gemini}. This comparison was done on a subset of the previously defined settings. We restricted this comparison to the two non-binary classification datasets (BIOS10 and Tweet Eval Emotion). Additionally, we only considered the positively fine-tuned versions of DistilBERT and T5, ensuring that all concept extraction methods were compatible. Finally, we used prompt types E1 and E2 and their anonymized variants and corresponding baselines to enable a consistent and fair comparison across different user-LLMs.

    % ----------------------------------------
    % User-LLMs Comparison Experiments Results
    \subsection{Comparison Experiments Results} \label{user_llms_res}

        \paragraph{Concept extraction methods.} The 3 user-LLMs give similar rankings in Tab.~\ref{tab:user_llm_comparison}; details in Fig.~\ref{appendix:fig: Gemini-1.5-flash_percentage} and Fig.~\ref{appendix:fig: Gemini-1.5-pro_percentage}. Not all pairwise differences are statistically significant for all user-LLMs as illustrated in Fig.~\ref{appendix:fig: Gemini-1.5-pro_difference} and Fig.~\ref{appendix:fig: Gemini-1.5-pro_difference} by the pairs NMF-\textit{NoProjection}, \textit{NoProjection}-SAE, and SAE-ICA. Nonetheless, it can be concluded that NMF ranks first in all settings, SAE and the \textit{NoProjection} baseline share the second place, pushing ICA to fourth. PCA and SVD are below both baselines.

        \paragraph{Concept interpretation methods.} Tab.~\ref{tab:user_llm_comparison} shows that the ranking is conserved across the different user-LLMs, placing CMAW on top. Differences are statistically significant.

% ========== %
% Conclusion %
% ========== %
% \section{Conclusion}
\section{Take-away}

    \paragraph{Summary.} We present a simulatability experiment for post-hoc unsupervised concept-based explanations with user-LLMs. The results show that concept-based explanations can help user-LLMs predict what would have predicted a classification model, and that user-LLM accuracy can be used to rank methods with statistical significance across multiple user-LLMs.

    \paragraph{Recommendations.} Our evaluation framework and the empirical report give concrete recommendations with regard to the different parts of concept-based explanations: the NMF method appears to be the most interpretable. However, it requires positive embeddings. Hence, without positive embeddings, we recommend the use of SAEs. These methods are popular in recent literature as they can create over-complete concept banks which are necessary for generative tasks. However, these models are fragile and difficult to implement. Thus, the ICA would be the simplest to apply as it does not have such constraints.
    
    Regarding the concept interpretability methods, using the CMAW only requires the model and has a constant cost, regardless of the number of methods or concepts. This makes it suitable as a baseline. It obtained better results than the second method. The method used by \cite{bricken2023monosemanticity,templeton2024scaling} seems to be more interpretable but requires much more computing, a more complex pipeline, and the use of an LLM.

    \paragraph{Discussions.} We argue that the positive concept space and sparse concept activations explain why the NMF and SAEs perform well. On their part, SVD and PCA condense the information across all samples in the fewest dimensions. This suggests that while sparsely activated for a given sample, the information should be distributed across all concept dimensions.
    
    Fig.~\ref{appendix: GPT-4o-mini_distribution} shows that user-LLMs do not grasp the model they must predict. We argue that the bottleneck of concept-based methods is the concept interpretation part. Suggesting that finding cost-efficient methods like CMAW and versatile methods as proposed by \citet{bricken2023monosemanticity} represent promising future work.

    On simulatability now, while it seems to evaluate a trade-off between faithfulness and complexity, determining its exact position remains an open question. Furthermore, depending on the meta-predictor, the evaluated explanation property may differ \cite{chan2022frame}.

% =========== %
% Limitations %
% =========== %
\section*{Limitations} \label{limitations}

    Despite our efforts to design thorough and comprehensive experiments, we acknowledge that certain blind spots and limitations may still remain, reflecting the inherent challenges in achieving complete coverage in such analyses. Namely, three major points could be raised:
    
    \begin{itemize}
        \item We followed the suggestion in~\cite{fel2023holistic} of computing the concepts in the penultimate layer of the model. We assume Computer Vision models behave similarly to NLP models in this regard, but this might not be the case. However, our framework can also be applied elsewhere in the residual stream or MLP layers of a transformer model.
        \item Due to the sudden popularity and speed at which the state-of-the-art of SAEs changes at the time of writing, the SAE studied in this work -- described in appendix \ref{appendix: sae} -- did not include the latest improvements \cite{rajamanoharan2024improving,rajamanoharan2024jumping,gao2024scaling,leask2024stitching,bussmann2024batchtopk}. Therefore, SAEs results are probably underestimated.
        \item Although previous work seems to provide evidence towards LLM's being a useful proxy for human behavior~\cite{de2024evaluating}, there is no actual proof that the ranking would be similar to one calculated using humans as meta-predictor $\Psi$.
    \end{itemize}

% =============== %
% Acknowledgments %
% =============== %
\section*{Acknowledgments}  
    
    This work was carried out within the DEEL project,\footnote{\url{https://www.deel.ai/}} which is part of IRT Saint Exupéry and the ANITI AI cluster. The authors acknowledge the financial support from DEEL's Industrial and Academic Members and the France 2030 program – Grant agreements n°ANR-10-AIRT-01 and n°ANR-23-IACL-0002.

    We would like to thank Thibaut Boissin for his help in making our figures clear and impactful.
    
    Finally, we extend a special thanks to reviewers from December 2024's ARR for suggesting the \textit{NoProjection} baseline and the comparison between automated simulatability and other metrics.

%\clearpage
\bibliography{biblio}

\appendix

% =================== %
% Experiment settings %
% =================== %
\section{Experiment Settings}

    \subsection{Datasets} \label{appendix:datasets}
    
        Some of the datasets were too small for concept extraction methods to converge to satisfying results. Therefore we artificially increased the datasets by adding modified versions of the samples. The new samples were obtained by splitting the initial ones by punctuation marks. Hence \texttt{[``This is a first example, made up for understanding.'']} becomes \texttt{[``This is a first example, made up for understanding.'', ``This is a first example'', ``made up for understanding'']}.

    % ------
    % Models
    \subsection{Models} \label{appendix:models}
    
        \paragraph{DistilBERT and T5.} The models used were extracted from HuggingFace. The model cards are: DistilBERT \cite{distilbertmodelcard}, T5 \cite{t5modelcard}, Llama-3-8B \cite{llama3modelcard}. For DistilBERT and T5, we used the \texttt{ModelForSequenceClassification} fine-tuned for each dataset.
        
        \paragraph{DistilBERT+ and T5+.} To build the positive versions, we added a ReLU function in the forward pass before the latent space we wanted to study. Then, these models were fine-tuned for the task.

        \paragraph{Llama.} We adapted \texttt{LlamaForCausalLM} to our task through prompting and only considered the next predicted token. The unembedding operation was used as our $g$ part of the model, and we limited it to the classes present in the dataset. The $h$ part was all the rest of the model.

    % -------------------
    % Concepts extraction
    \subsection{Concepts Extraction Methods} \label{appendix:methods}
        
        The goal is to define a concept space $\mathcal{C} \subseteq \mathbb{R}^k$ through the concept transformation $t: \mathcal{H} \longrightarrow \mathcal{C}$ and its bijection (or approximation) $t^{-1}: \mathcal{C} \longrightarrow \mathcal{H}$. We note $X \in \mathcal{X}^n$ a set of samples, $A \in \mathcal{H}^n$ their latent embeddings, and $U \in \mathcal{C}^n$ their projection in the concept space. Similarly, respectively, we note $x$, $a$, and $u$ as elements of these sets.
        Unlike many unsupervised concept-based explicability methods, here we make \textbf{a single projection} for the task and not a projection for each predicted class (similar to \citep{jourdan2023taco}).

        Apart from SAE, all implementations are from scikit-learn \cite{scikit-learn} with default parameters apart from the `n\_components` that we vary with the number of required concepts. The used classes are \texttt{FastICA}, \texttt{NMF}, \texttt{PCA}, and \texttt{TruncatedSVD}.

        % NoProjection
        \paragraph{NoProjection} considers the latent space as the concept space. Hence, latent space neurons correspond to concepts. Such a paradigm was studied in previous work \cite{geva2022transformer}. Nonetheless, we consider it as a baseline.

        % ICA
        \paragraph{Independent Component Analysis (ICA)} \cite{ans1985architectures,hyvarinen2000independent} extracts independent components or sources $S$ such that $S = W \cdot whiten(A)$. The whitening function centers the data on 0. We could write $whiten(a) = a - \mu$. Therefore, in our case we could define $t_{ICA}(a):= W \cdot (a - \mu)$. Then we can compute the Moore-Penrose pseudo-inverse $W^+$, hence we can define $t^{-1}_{ICA}(u):= W^+u + \mu$.
        
        We use the \texttt{FastICA} implementation from scikit-learn \cite{scikit-learn} with default parameters apart from the `n\_components` that we vary with the number of required concepts.

        % NMF
        \paragraph{Non-negative Matrix Factorization (NMF)} \cite{lee1999learning,sra2005generalized} factorizes the matrix $A$ into two matrices $U$ and $W$ such that $A = UW$. The particularity of the NMF is that all three matrices have non-negative weights.

        It is easy to construct $t^{-1}$ as $U$ is the concepts activations, thus $t^{-1}(u) = uW$. But this factorization is nonlinear, and we cannot inverse $W$. Therefore, to obtain $U_2$ corresponding to other latent embeddings $A_2$, an $U_2$ is optimized to fit $A_2 = U_2W$ with W fixed. Hence, $t$ cannot be defined by matrix multiplications and can only be done by solving an equation.

        We use the \texttt{NMF} implementation from scikit-learn \cite{scikit-learn} with default parameters apart from the `n\_components` that we vary with the number of required concepts.

        % PCA
        \paragraph{Principal Component Analysis (PCA)} \cite{pearson1901liii,hotelling1992relations} transforms a zero-centered matrix $A - \mu$ into another matrix $U$ through linear combinations $W$ such that $U = (A- \mu)W$. Hence we can define $t$ by $t(a) = (a - \mu)W$ once $W$ is computed, then by investing $W$ we define $t^{-1}(u) = uW^{-1} + \mu$.

        We use the \texttt{PCA} implementation from scikit-learn \cite{scikit-learn} with default parameters apart from the `n\_components` that we vary with the number of required concepts.

        % Sparse Auto-Encoder
        \paragraph{Sparse Auto-Encoder (SAE)} \label{appendix: sae} \cite{ng2011sparse,makhzani2013k,domingos2015master} are neural networks whose outputs should be the same as the inputs, the particularity is that some constraints are applied in the middle during their training. Hence $t$ is the encoder and $t^{-1}$ the decoder.
        
        In our case, we follow most of the recommendations from \citet{bricken2023monosemanticity}. In some, we use a $\ell1$ component with a $1e-3$ coefficient in the loss to push toward sparsity. We apply dead neuron resampling; this part is very sensitive to hyperparameter modifications. Finally, we do $100,000$ steps with a learning rate starting at $1e-3$. Note that in some cases, early stopping fires.

        % SVD
        \paragraph{Singular Value Decomposition (SVD)} \cite{eckart1936approximation} factorizes the matrix $A$ into three components, such that $A = U \Sigma V^T$. In our case, we use $U\Sigma$ as concept activations we note $U$. Hence, with our notations, we have $A = UV^T$. Since with the SVD, $V^TV = I$, then we can define the projections by $t(a) = aV$ and $t^{-1}(u) = uV^T$.

% ======================== %
% Simulatability Prompting %
% ======================== %
\section{Simulatability Prompting}

    % --------------
    % Prompt samples
    \subsection{Prompt Samples} \label{appendix: prompt_samples}

        In our simulatability experiments, we select $40$ samples for each dataset-model pair: $20$ for the Learning Phase (LP) and $20$ for the Evaluation Phase (EP). These samples are chosen to cover each class uniformly. Among them, $20$ are correctly classified by $f$, and $20$ are misclassified, ensuring a balanced challenge for the meta-predictor. We then randomly distribute these samples between the LP and EP.

        To increase statistical robustness, we repeat this selection with $5$ different random seeds, resulting in $5$ distinct sets of 40 samples. Additionally, we use 2 more sets of 40 samples to determine the optimal number of concepts for each dataset-method pair.

        Finally, this paper reports $23,360$ for GPT-4o-mini and $960$ for both Gemini-1.5 Flash and Pro. These prompts had a mean number of tokens around $2,000$, mostly represented by the samples.

    % ------------------------------
    % Normalizing concept importance
    \subsection{Normalizing Concept Importance} \label{appendix: normalize_importance}

        For a given class $c$ and concept $cpt$, its normalized global importance is defined by:

        \begin{equation}
            \hat{\varPhi}_{f_{co},c,cpt} = \frac{\varPhi_{f_{co},c,cpt}}{\sum_{i=1}^{k}|\varPhi_{f_{co},c,i}|}
            \label{appendix:eq:normalized_global_importance}
        \end{equation}

        For a local explanation of sample $x$ with concepts projection $u = f_{ic}(x)$, the normalized local importance for a given concept $cpt$ is given by:

        \begin{equation}
            \hat{\varphi}_{f_{co}(u)_{cpt}} = \frac{\varphi_{f_{co}(u)_{cpt}}}{\sum_{i=1}^{k}|\varphi_{f_{co}(u)_{i}}|}
            \label{appendix:eq:normalized_local_importance}
        \end{equation}

    % ---------------------------------
    % Communicate importance in prompts
    \subsection{Communicate Importance in Prompts} \label{appendix:communicate_importance}

        In several cases, communicating how important some concepts are is necessary. However, LLMs have been proven to be unable to compare numerical values. Furthermore, encoding importance value via a single token would make the comparison far easier. Finding a one-token word to encode these values was not trivial. Hence we opted for the signs "- -", "-", "+", and "+ +". Concepts with low local importance were not shown for local explanations either. To decide what sign to show, we used arbitrary thresholds. The correspondences can be found in Tab.~\ref{appendix:tab:importance_encoding}. Nonetheless, we evaluated other important communication solutions in preliminary experiments, such as numerical values and characters such as "high" and "low". The signs were the best-performing solution.

        \begin{table}[h]
            \centering
            \ra{1.2}
            \scriptsize
            \begin{tabular}{@{}ccccc@{}}
                \toprule
                $\hat{\varphi}$ intervals   & [-1, -0.3] & ]-0.3, -0.05] & [0.05, 0.3[ & [0.3, 1] \\
                \midrule
                Encoding                    & "- -"       & "-"           & "+"         & "+ +"     \\
                \bottomrule
            \end{tabular}
            \caption{Table of buckets for concept importance encoding in prompts.}
            \label{appendix:tab:importance_encoding}
        \end{table}

% ======================================== %
% Experiments Supplementary Visualizations %
% ======================================== %
\section{Experiments Supplementary Visualizations}

    % --------------------------------
    % Extended GPT-4o-mini experiments
    \subsection{Ranking Methods with GPT-4o-mini}  \label{appendix: GPT-4o-mini_res}

        % \begin{figure*}[ht]
        %     \centering
        %     \includegraphics[width=\textwidth]{plots/gpt-4o-mini_extended_E_violin.pdf}
        %     \caption{Violin plot of simulatability scores distribution by prompt type and concept extraction method. Samples used were selected to be problematic. The three right prompt types are on anonymous classes. Experiments from Sec.~\ref{GPT-4o-mini_xp}.}
        %     \label{appendix:fig: GPT-4o-mini_violin}
        % \end{figure*}

        % Fig.~\ref{appendix:fig: GPT-4o-mini_violin} illustrates simulatability scores distributions via a violin plots. For prompt types with clear class names, the methods are all equivalent to the baseline. This means that the meta-predictor $\Phi$ shortcuts the task; it tries to predict the initial labels and not the predictions from the model $f$.

\begin{figure}[t]
            \centering
            \includegraphics[width=0.48\textwidth]{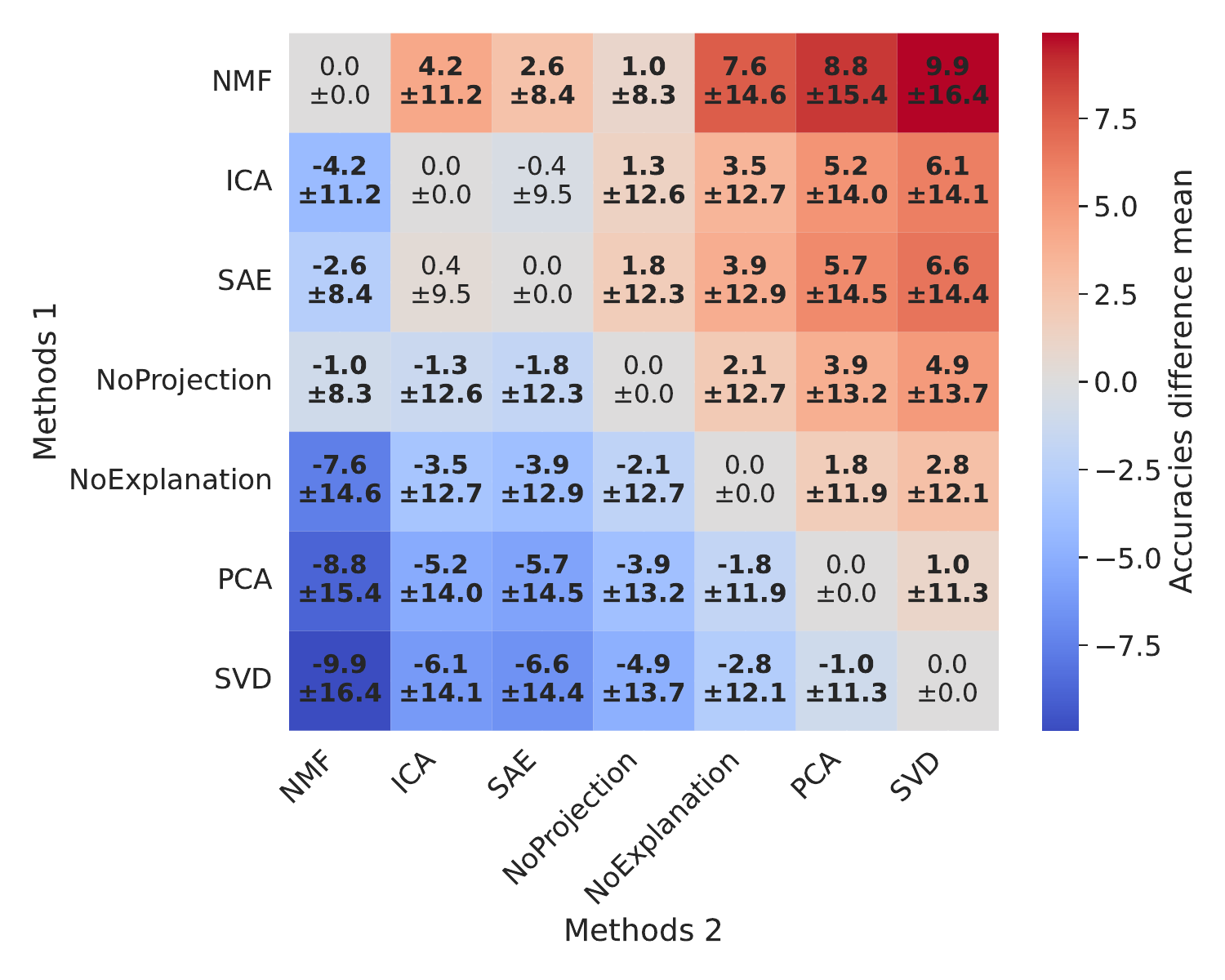}
            \caption{Pairwise comparison matrices on GPT-4o-mini experiments described in Sec.~\ref{GPT-4o-mini_xp}. Difference means and standard deviations between method 1 and method 2 simulatability scores across experiments. Bold differences are statistically significant.}
            \label{appendix:fig: GPT-4o-mini_difference}
        \end{figure}

        Fig.~\ref{fig: GPT-4o-mini_percentage} and Fig.~\ref{appendix:fig: GPT-4o-mini_difference} illustrate concept extraction methods pairwise comparison matrices when all GPT-4o-mini settings are taken into account.

\begin{figure*}[t]
            \centering
            \includegraphics[width=0.70\textwidth]{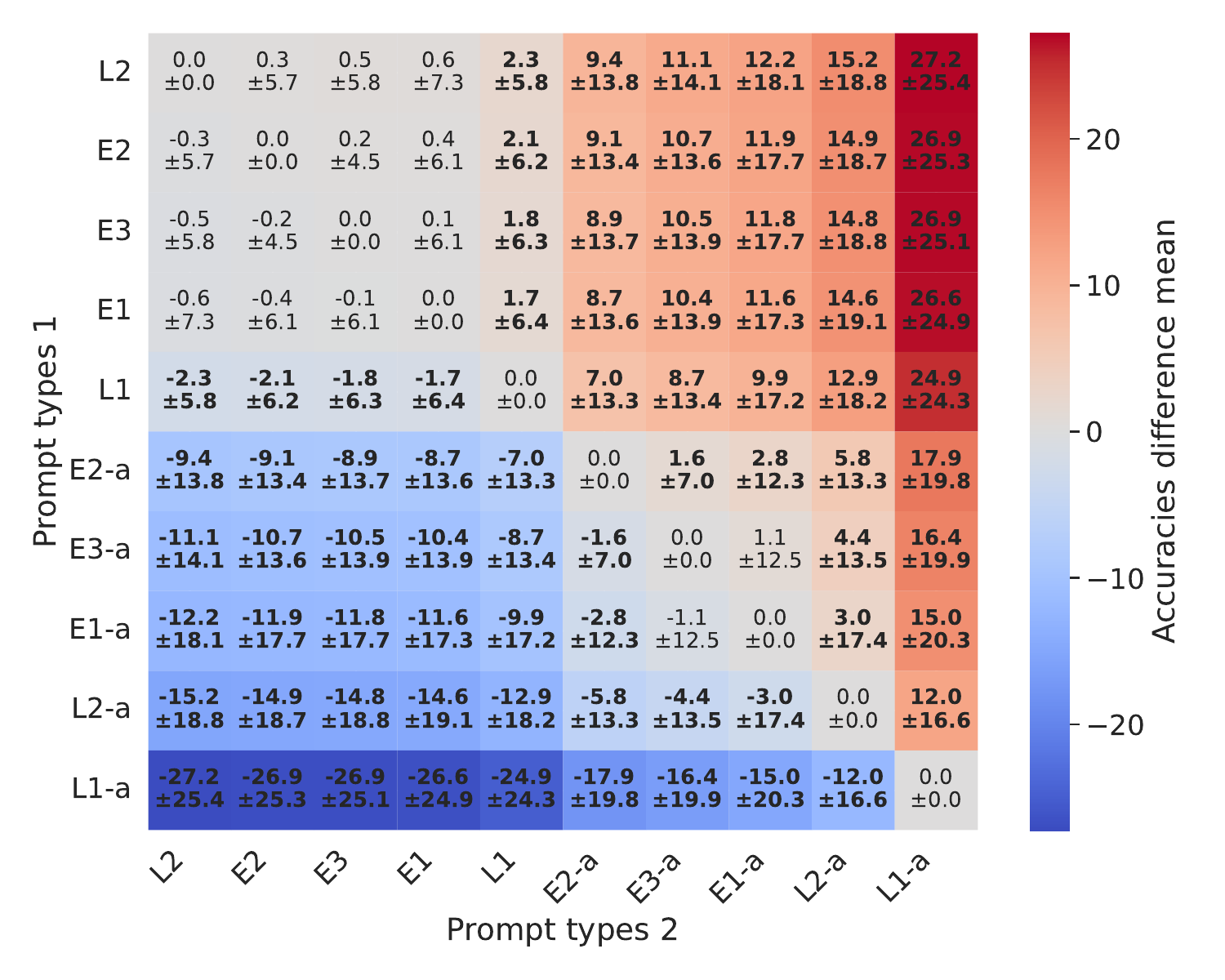}
            \caption{Percentage of simulatability experiments where method 1 is over method 2. Ranking by number of pairwise victories. GPT-4o-mini experiments described in Sec.~\ref{GPT-4o-mini_xp} subset to the top 3 methods NMF, SAE, and ICA. Experiments (E1, E2, and E3) and baselines (NE1 and NE2) are described in Sec.~\ref{prompt_types} and Tab.~\ref{tab:prompt_types}. They differ in the simulatability elements present in the prompt. Experiments and baselines with "-a" are done with anonymous classes.}
            \label{appendix:fig: GPT-4o-mini_prompt_types}
        \end{figure*}

        Fig.~\ref{appendix:fig: GPT-4o-mini_prompt_types} shows the pairwise comparison matrix with the percentages of wins between prompt types. It takes into the top3 concept extraction methods (NMF, SAE, and ICA) and all of the GPT-4o-mini settings on other variables.

        Tab.~\ref{appendix:tab: GPT-4o-mini_res_methods} and Tab.~\ref{appendix:tab: GPT-4o-mini_res_methods} show the rankings of concept extraction methods and concept interpretation methods when the other is fixed. The order is conserved in both cases, and the NMF-CMAW pair emerged in the first rank.

        \begin{table}[t]
            \centering
            \scriptsize
            \ra{1.2}
            \begin{tabular}{@{}lccccccc@{}}
                \toprule
                \tb{Method} & \tb{NMF} & \tb{SAE} & \tb{ICA} & \tb{NoPr} & \tb{NoEx} & \tb{PCA} & \tb{SVD} \\ \midrule
                CMAW        & \tb{1}   &     3    & \ul{2}   &     4     &    5      &     6     &    7     \\
                o1CA        & \tb{1}   & \tb{1}   &     4    & \tb{1}    &    5      &     6     &    7     \\
                \bottomrule
            \end{tabular}
            \caption{\textbf{Experiments with GPT-4o-mini as a user-LLM.} Concepts extraction methods ranking for the two concept interpretation methods.}
            \label{appendix:tab: GPT-4o-mini_res_methods}
        \end{table}

        \begin{table}[t]
            \centering
            \ra{1.2}
            \scriptsize
            \begin{tabular}{@{}lccc@{}}
                \toprule
                Extraction   & \textbf{CMAW} & \textbf{o1CA} & \textbf{NoExplanation} \\
                \midrule
                NMF          & \tb{1}        & \ul{2}        &     3     \\
                SAE          & \tb{1}        & \ul{2}        &     3     \\
                ICA          & \tb{1}        & \ul{2}        &     3     \\
                NoProjection & \tb{1}        & \ul{2}        &     3     \\
                \bottomrule
            \end{tabular}
            \caption{\textbf{Experiments with GPT-4o-mini as a user-LLM.} Concepts interpretation methods ranking for the top 3 concept extraction methods and the NoProjection baseline. The order is the same for all methods.}
            \label{appendix:tab: GPT-4o-mini_res_interpretability}
        \end{table}

    % ----------------------------------------------
    % GPT-4o-mini simulatability scores distribution
    \subsection{GPT-4o-mini simulatability scores distribution} \label{appendix: GPT-4o-mini_distribution}

        \begin{figure*}
            \centering
            \begin{subfigure}{\textwidth}
                \centering
                \includegraphics[width=\textwidth]{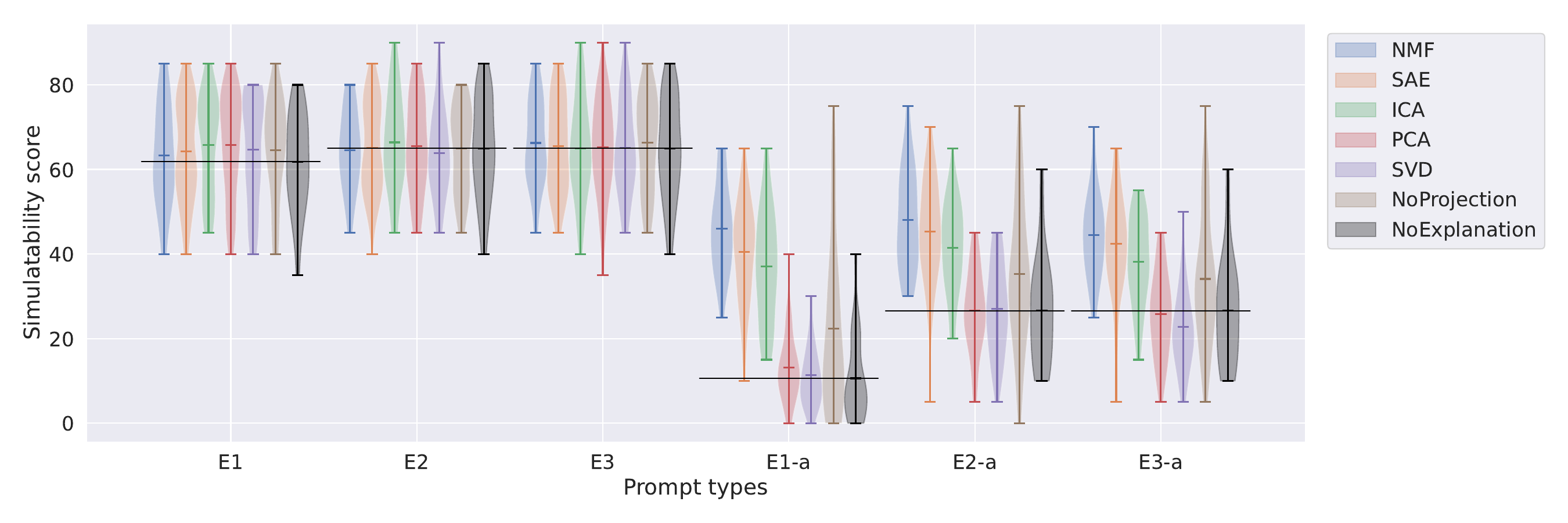}
                \caption{BIOS10 (10 classes)}
                \label{fig: BIOS10_distribution}
            \end{subfigure}
            
            \begin{subfigure}{\textwidth}
                \centering
                \includegraphics[width=\textwidth]{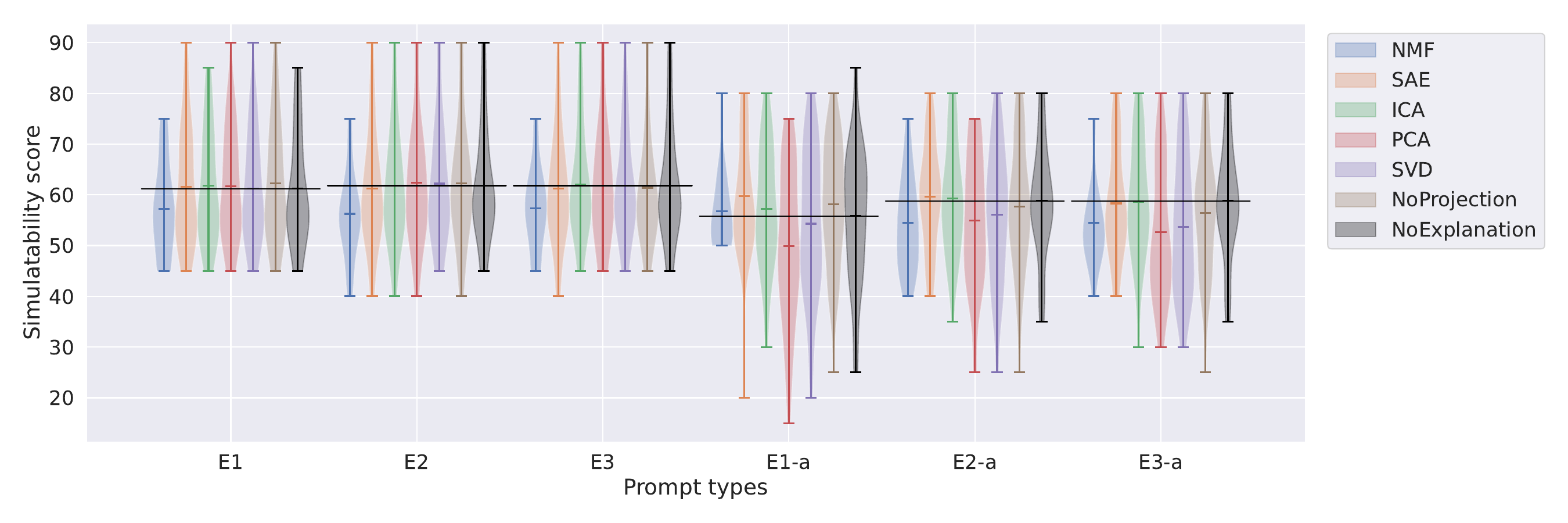}
                \caption{IMDB (2 classes)}
                \label{fig: IMDB_distribution}
            \end{subfigure}
            
            \begin{subfigure}{\textwidth}
                \centering
                \includegraphics[width=\textwidth]{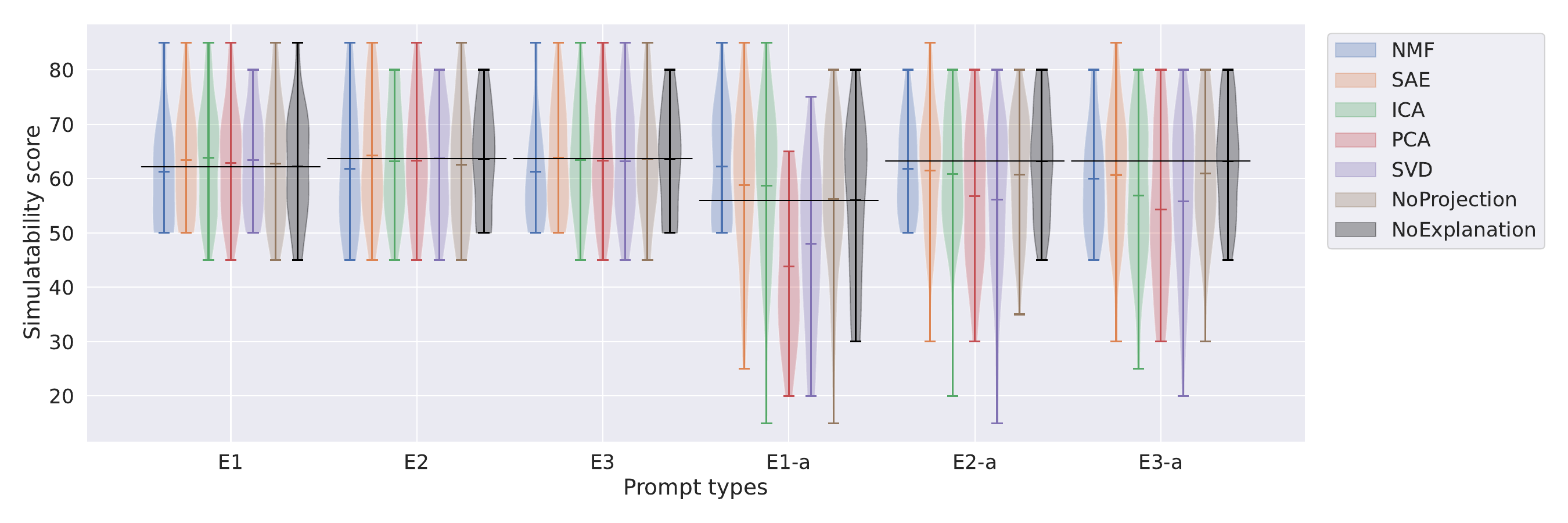}
                \caption{Rotten Tomatoes (2 classes)}
                \label{fig: rotten_tomatoes_distribution}
            \end{subfigure}
        
            \begin{subfigure}{\textwidth}
                \centering
                \includegraphics[width=\textwidth]{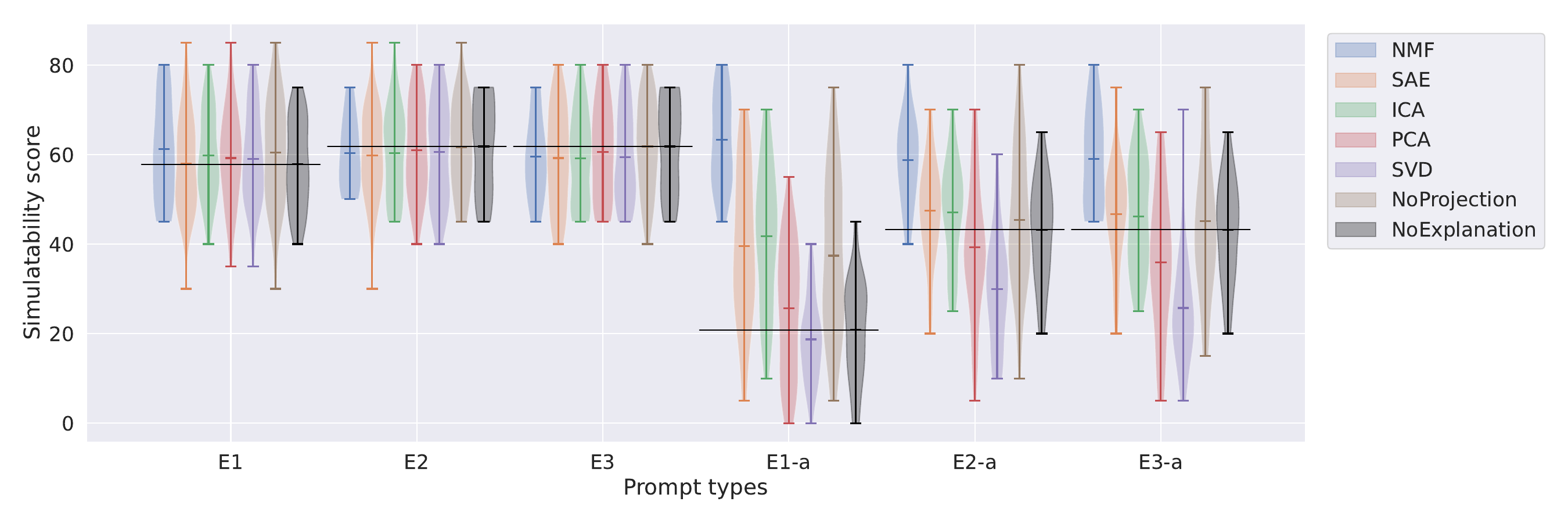}
                \caption{Tweet Eval Emotion (4 classes)}
                \label{fig: tweet_eval_emotion_distribution}
            \end{subfigure}
            
            \caption{Distribution of simulatability scores for each concept-extraction method and prompt type pairs. Simulatability scores are accuracies between the initial model $f$ predictions and the meta-predictor $\Psi$ predictions. Studied samples are representative and diverse, with half of them having wrong initial model predictions with regard to the labels. Plots are divided between datasets as the accuracy highly depends on the number of classes. Here, the methods only improve significantly over the baseline for the anonymous classes prompt types.}
            \label{fig: appendix: GPT-4o-mini_distribution}
        \end{figure*}

        Fig.~\ref{appendix: GPT-4o-mini_distribution} represents the simulatability scores distributions on the different datasets for each concept-extraction method and prompt type pairs. It shows that the methods and the baselines obtain similar results without anonymous classes, suggesting that the meta-predictor shortcuts the tasks. However, with the anonymous classes experiments (E1-a, E2-a, and E3-a), some methods significantly outperform the baselines.

    % --------------------
    % User-LLMs comparison
    \subsection{User-LLMs Comparison} \label{appendix:user_llms_res}

        \begin{figure*}[t]
            \centering
            \begin{subfigure}[b]{0.49\textwidth}
                \includegraphics[width=\textwidth]{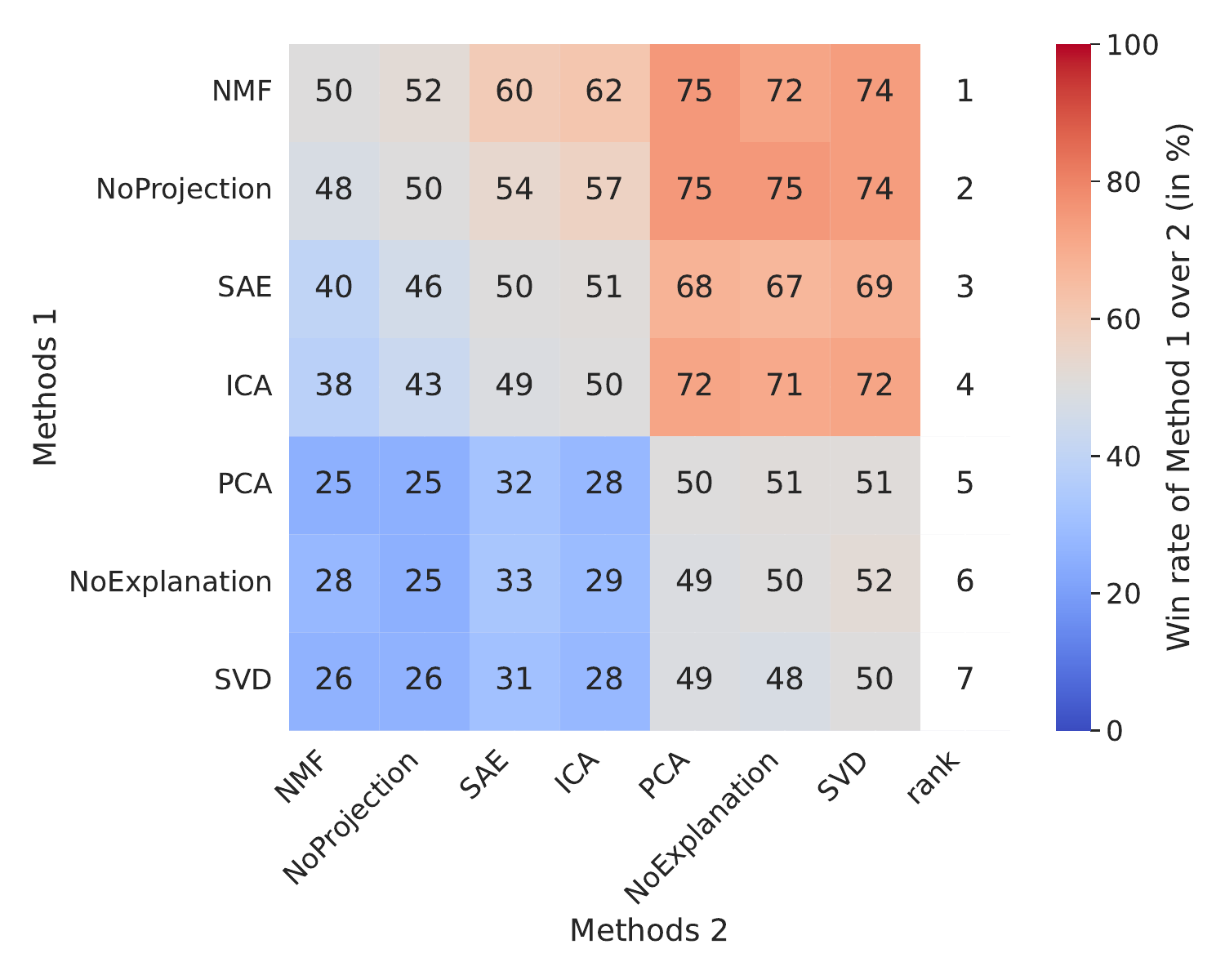}
                \caption{Percentage of simulatability experiments where method 1 is over method 2. Ranking by number of pairwise victories.}
                \label{appendix:fig: Gemini-1.5-flash_percentage}
            \end{subfigure}
            \begin{subfigure}[b]{0.49\textwidth}
                \includegraphics[width=\textwidth]{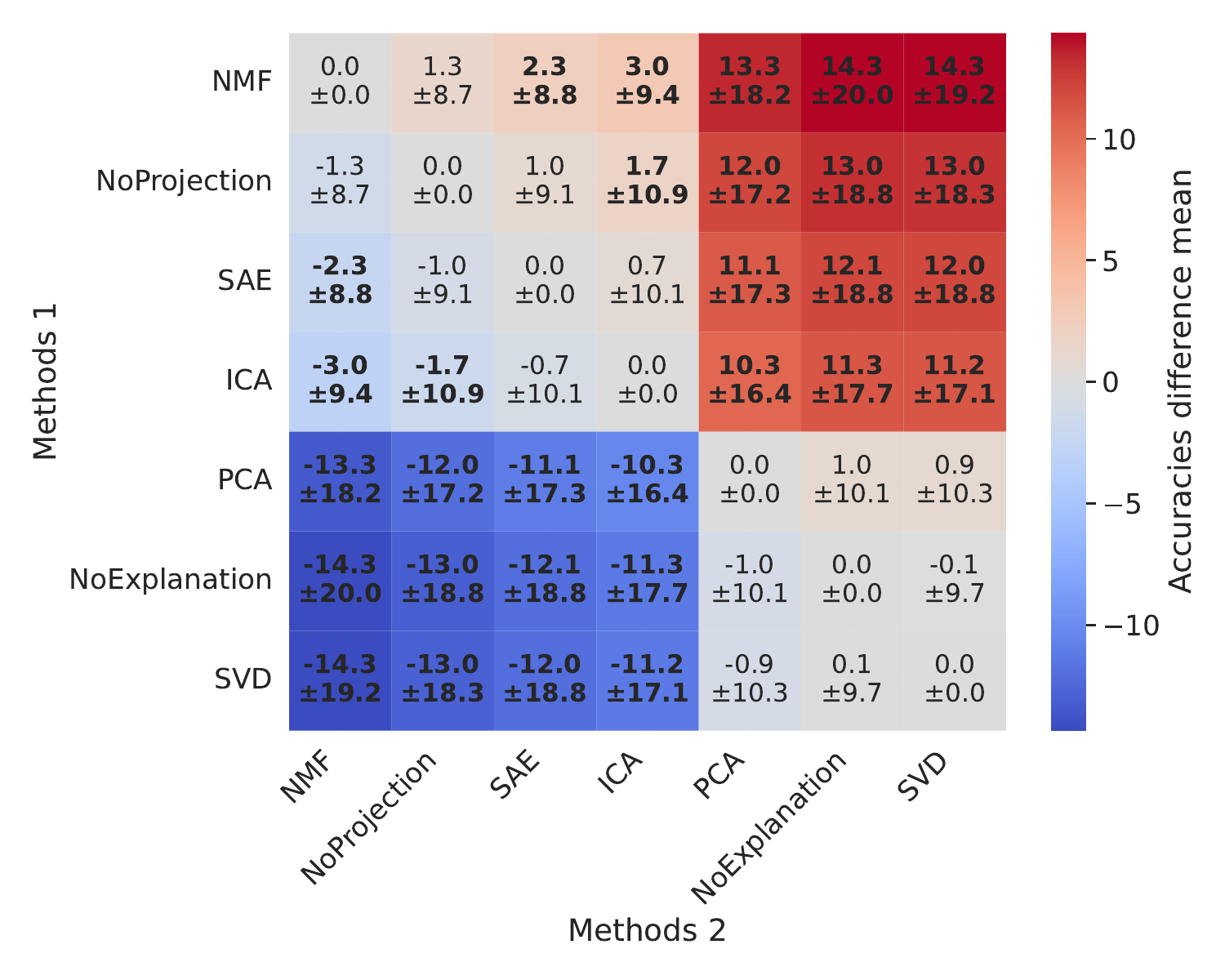}
                \caption{Difference means and standard deviations between method 1 and method 2 simulatability scores across experiments. Bold differences are statistically significant.}
                \label{appendix:fig: Gemini-1.5-flash_difference}
            \end{subfigure}
            \caption{Pairwise comparison matrices on Gemini-1.5-Flash experiments described in Sec.~\ref{user_llms_xp}. NMF comes first, and with SAE and ICA, these methods significantly improve over the baseline (\textit{i.e.} without explanations).}
            \label{appendix:fig: Gemini-1.5-flash_res}
        \end{figure*}

        Fig.~\ref{appendix:fig: Gemini-1.5-flash_res} illustrates concept extraction methods, pairwise comparison matrices for Gemini-1.5-flash as the meta-predictor.

        \begin{figure*}[t]
            \centering
            \begin{subfigure}[b]{0.49\textwidth}
                \includegraphics[width=\textwidth]{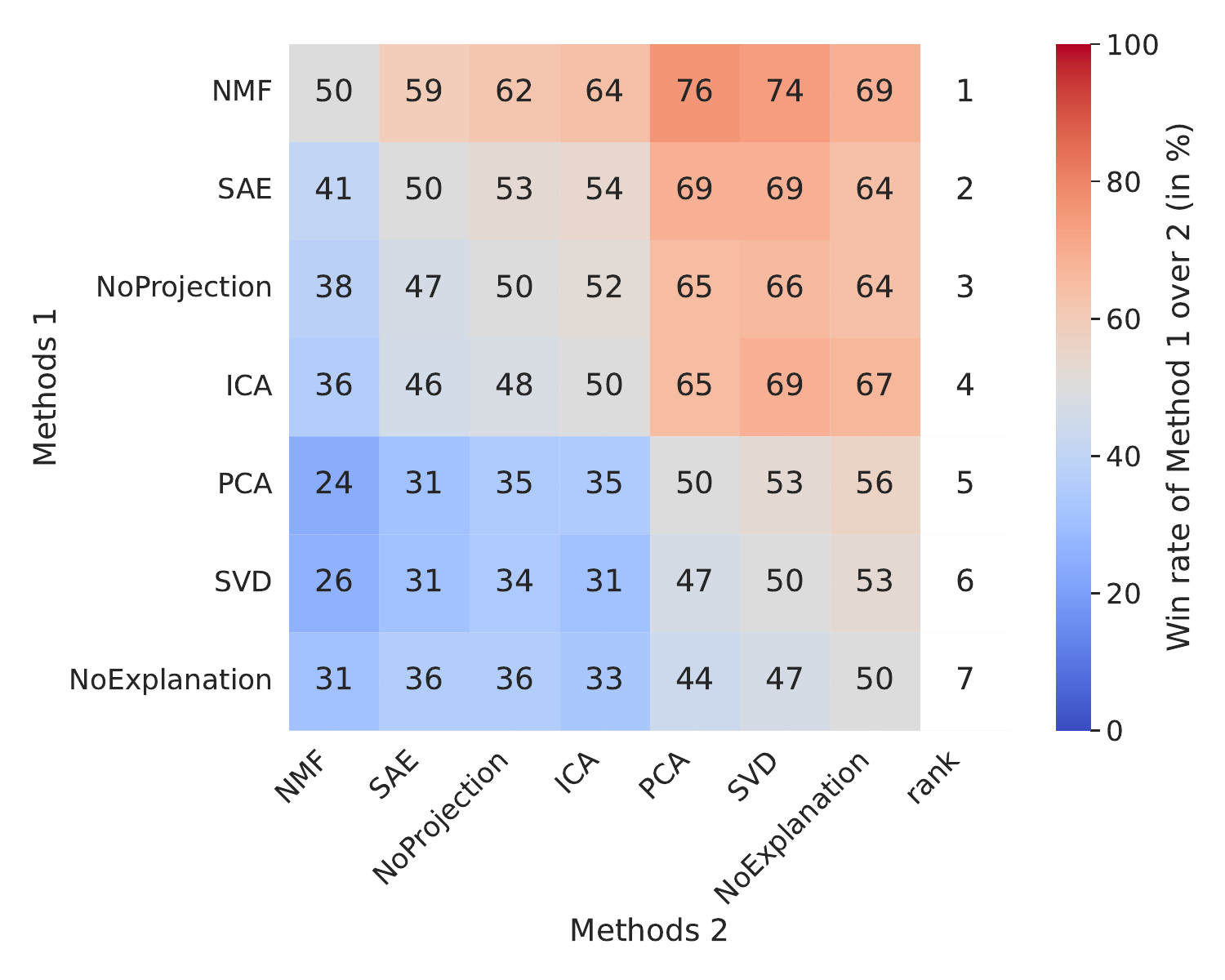}
                \caption{Percentage of simulatability experiments where method 1 is over method 2. Ranking by number of pairwise victories.}
                \label{appendix:fig: Gemini-1.5-pro_percentage}
            \end{subfigure}
            \begin{subfigure}[b]{0.49\textwidth}
                \includegraphics[width=\textwidth]{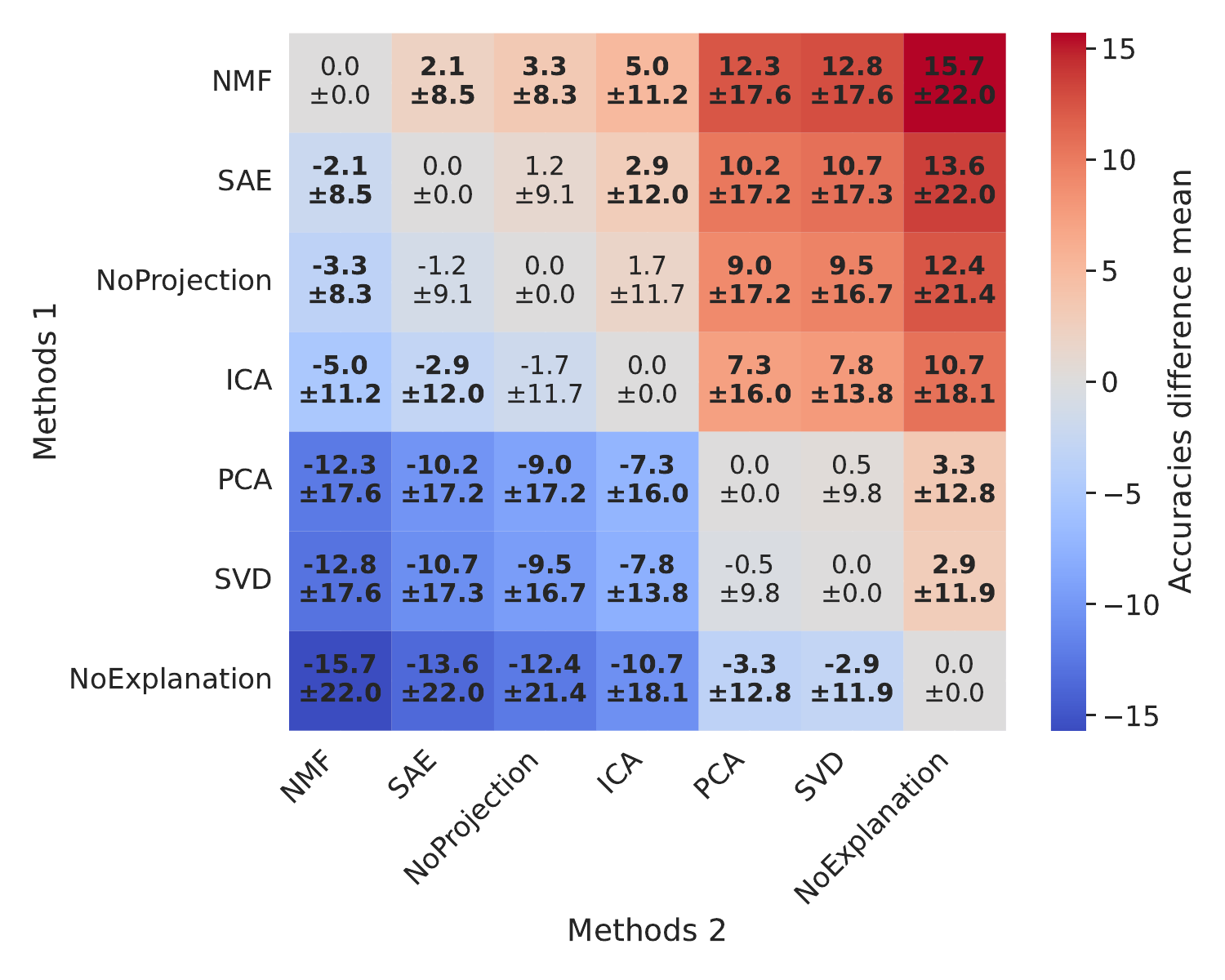}
                \caption{Difference means and standard deviations between method 1 and method 2 simulatability scores across experiments. Bold differences are statistically significant.}
                \label{appendix:fig: Gemini-1.5-pro_difference}
            \end{subfigure}
            \caption{Pairwise comparison matrices on Gemini-1.5-Pro experiments described in Sec.~\ref{user_llms_xp}. NMF comes first, and with SAE and ICA, these methods significantly improve over the baseline (\textit{i.e.} without explanations).}
            \label{appendix:fig: Gemini-1.5-pro_res}
        \end{figure*}

        Fig.~\ref{appendix:fig: Gemini-1.5-pro_res} illustrates concept extraction methods, pairwise comparison matrices for Gemini-1.5-pro as the meta-predictor.

% ============= %
% Other Metrics %
% ============= %
\section{Other Metrics} \label{appendix: other metrics}

    Eleven metrics were compared to simulatability via Spearman's rank correlation \cite{spearman1904proof} in Fig.~\ref{fig: metrics correlations}. This section describes the different metrics evaluated corresponding to the name we used in our visualization. These metrics evaluate either complexity or faithfulness.

    % ----------
    % Complexity
    \subsection{Complexity}

        \paragraph{nb concepts.} Represent the number of dimensions of the concept space of evaluated methods. Also denoted $k$ such that $\mathcal{C} \subseteq \mathbb{R}^k$.

        \paragraph{nb activated or L0.} It is the mean number of concepts with non-zero values across the dataset $X$.
        $$L0_X = \expectation\limits_{x \in X} \sum_{i=0}^k\mathds{1}\left\{f_{ic}(x)_i \neq 0\right\}$$

        \paragraph{ratio of activated.} It is the number of activated concepts over the total number of concepts. We could call it sparsity. It could be written $\frac{L0_X}{k}$.

        \paragraph{cosine similarity.} The cosine similarity of a concept extraction method is a measure of how similar are the different concepts of a method. It is computed on the concept decoder weight $W_{t^{-1}}$ assuming that $t^{-1}(u) = uW_{t^{-1}}$. The metric can thus be written:

        $$\frac{1}{k^2} \sum_{i,j}^k cosine(W_{t^{-1}, i}, W_{t^{-1}, j})$$

        \paragraph{covariance.} Similarly to the cosine similarity, the covariance is also computed between the concepts of a single method. It is the mean value of the covariance matrix between concepts. Computing the correlation between concepts resulted in the exact same metric.

        \paragraph{nb important.} This corresponds to the number of concepts shown in the prompts. The number of important concepts depends on the concept important method used. Here, because we split the model on the penultimate layer, input-gradient was an optimal method. Therefore, we still use it for the metrics. We decided to use a $5\%$ threshold, such that for each class $c$, concepts with $\hat{\varPhi}_{f_{co},c,cpt}$ \ref{appendix:eq:normalized_global_importance} over this threshold $\tau$ were considered important:

        $$\left|\bigcup_c\left\{cpt \in [1,k] | \hat{\varPhi}_{f_{co},c,cpt} > \tau \right\}\right|$$

        \paragraph{ratio of important.} The ratio of important concepts corresponds to the number of important concepts over the total number of concepts. It can also be seen as the percentage of concepts shown in the prompts.

    % ------------
    % Faithfulness
    \subsection{Faithfulness}

        \paragraph{latents L2.} The latent embeddings $L2$ reconstruction error for a set of samples $X$ is computed between the initial latent embeddings and the reconstructed latent embeddings:

        $$||h(x) - t^-1(f_{ic}(x))||_{x \in X}^2$$

        \paragraph{logits L2.} The logits $L2$ reconstruction error is computed similarly to the latent embeddings reconstruction error:

        $$||f(x) - f_c(x)||_{x \in X}^2$$

        \paragraph{logits KL.} The Kullback-Leibler divergence \cite{kullback1951information} compares the logits distribution between $f$ and $f_c$ on the set of samples $X$:

        \paragraph{completeness.} The completeness score was defined by \citet{yeh2020completeness}, we follow their formula.

    % Fig.~\ref{appendix:fig: metrics correlations} illustrate Spearman's rank correlation \cite{spearman1904proof} between simulatability and eleven other metrics. It shows that higher faithfulness positively correlates with simulatability. Also, the more concentrated the information is, in terms of the number of activated concepts, the higher the simulatability is. This could explain why NMF and SAE are the two best-performing methods. They are the only method with a ratio of activated concepts lower than $1$.

    % \begin{figure*}[t]
    %     \centering
    %     \includegraphics[width=\linewidth]{plots/GPT-4o-mini_metrics_correlation.pdf}
    %     \caption{Comparison between simulatability and other concept-based metrics. Other metrics are divided between faithfulness and complexity. It shows that higher fidelity and lower complexity tend to increase simulatability.}
    %     \label{appendix:fig: metrics correlations}
    % \end{figure*}

% ============== %
% Prompt example %
% ============== %
\section{Prompt example} \label{appendix:prompt_example}

    This appendix shows an example of a prompt for the BIOS10 dataset with the T5+ model. The concept extraction method used was the NMF with 20 concepts. The communication method was the first one, with the 5 most activating words. All the other prompt types could be constructed from this one by removing elements following Tab.~\ref{tab:prompt_types}. The only other thing to modify would be the task description.
    \bigskip

    The 40 samples were reduced to 20, 10 for LP and 10 for EP (only in the visualization for the paper). Otherwise, it would take up too much space. Furthermore, long biographies were cut for visualization purposes in the paper. All elements can be found in the 7 following colored boxes.
    % listings \ref{appendix:prompt_task_description}, \ref{appendix:prompt_global_explanations}, \ref{appendix:prompt_lp_samples}, \ref{appendix:prompt_lp_local_explanations}, \ref{appendix:prompt_lp_predictions}, \ref{appendix:prompt_ep_samples}, \ref{appendix:prompt_ep_local_explanations}, and \ref{appendix:prompt_responses}.

    % --------------------
    % IP: Task description
        \begin{figure*}[t]
        \begin{tcolorbox}[colframe=black!60,colback=white,title=1.1 IP: Task description | \textit{In all prompt types, but adapts to the following elements.}]
            You are a classifier. For each sample, you have to predict the class. To complete the task, you will be given the concepts and their importance for each class. You will have examples of samples, labels, and concepts' contributions to labels as references for the task. Each sample class prediction should be in the format: 'Sample\_\{i\}: \{predicted\_class\}'.\\
        
            The classes are: [surgeon, photographer, professor, teacher, physician, journalist, attorney, nurse, dentist, psychologist]
        \end{tcolorbox}
        \end{figure*}

    % ----------------------
    % IP: Global explanation
    \begin{figure*}[t]
    \begin{tcolorbox}[colframe=black!60, colback=white, title=1.2 IP: Global explanations | \textit{E1, E2, and E3}]
    For each concept, the most aligned and opposed words are:\\
    concept\_1: aligned: [''', 'paris', 'mit', 'france', 'vincent']\\
    concept\_3: aligned: ['faith', 'elizabeth', 'henry', 'kelly', 'london']\\
    concept\_4: aligned: ['attorneys', 'attorney', 'lawyers', 'lawyer', 'counsel']\\
    concept\_5: aligned: ['nurses', 'nurse', 'nursing', 'emergency', 'obstetrics']\\
    concept\_7: aligned: ['photography', 'photographer', 'shoots', 'photographers', 'portraits']\\
    concept\_8: aligned: ['psychotherapy', 'psychologist', 'psychology', 'counseling'...\\  %, 'psychological']\\
    concept\_9: aligned: ['surgeon', 'neurosurgery', 'surgery', 'surgeons', 'surgical']\\
    concept\_10: aligned: ['teachers', 'teacher', 'classroom', 'pedagogy', 'teaching']\\
    concept\_11: aligned: ['journalists', 'journalism', 'journalist', 'reporter', 'news']\\
    concept\_12: aligned: ['dentistry', 'dentist', 'dental', 'tooth', 'teeth']\\
    concept\_13: aligned: ['resolution', 'perception', 'globalization', 'welcome', 'impaction']\\
    concept\_14: aligned: ['patel', 'bangalore', 'delhi', 'nagar', 'india']\\
    concept\_15: aligned: ['immunology', 'pathology', 'cardiology', 'radiology', 'epidemiology']\\
    concept\_17: aligned: ['medicine', 'metabolism', 'healthcare', 'doctors', 'arab']\\
    concept\_19: aligned: ['professor', 'scholarship', 'scholars', 'interdisciplinary', 'co-editor']\\
    \\
    The most important concepts and their importance for each class are:\\
    surgeon: {'concept\_1': '+', 'concept\_9': '+'}\\
    photographer: {'concept\_7': '+'}\\
    professor: {'concept\_13': '+', 'concept\_15': '+', 'concept\_19': '+'}\\
    teacher: {'concept\_3': '+', 'concept\_10': '+'}\\
    physician: {'concept\_1': '+', 'concept\_15': '+', 'concept\_17': '+'}\\
    journalist: {'concept\_11': '+'}\\
    attorney: {'concept\_4': '+'}\\
    nurse: {'concept\_5': '+'}\\
    dentist: {'concept\_12': '+', 'concept\_14': '+'}\\
    psychologist: {'concept\_8': '+'}
    \end{tcolorbox}
    \end{figure*}

    % -----------
    % LP: Samples
    \begin{figure*}[t]
    \begin{tcolorbox}[colframe=black!60,colback=white,title=2.1 LP: Samples | \textit{NE2, E2, and E3}] 
    Sample\_0:  He is professionally affiliated with Dartmouth-Hitchcock Medical...\\
    Sample\_1:  Unfortunately he has a genetic condition called Alport syndrome that...\\
    Sample\_2:  Working closely with ENT physicians, she not only cares for these...\\
    Sample\_3:  She practices in Apo, Armed Forces Europe and has the professional...\\
    Sample\_4:  He is also a senior economics commentator for the Guardian, where he...\\  
    Sample\_5:  Sarah specializes in bringing people together to design products...\\
    Sample\_6:  He is especially interested in dental procedures. He is professionally...\\ 
    Sample\_7:  She is also a Clinical Psychologist and a Senior Researcher at the...\\  
    Sample\_8:  Some of her favorite people include those who step out of her thoughts...\\
    Sample\_9:  He's also a garage mechanic who bought cars cheap in high school...
    \end{tcolorbox}
    \end{figure*}

    % ---------------------
    % LP: Local explanation
    \begin{figure*}[t]
    \begin{tcolorbox}[colframe=black!60,colback=white,title=2.2 LP: Local explanation | \textit{Only in E3}]
    Concepts contributions for Sample\_0: {'concept\_1': '+', 'concept\_17': '+'}\\
    Concepts contributions for Sample\_1: {'concept\_1': '-', 'concept\_3': '+', 'concept\_9': '-', 'concept\_11': '+'}\\
    Concepts contributions for Sample\_2: {'concept\_3': '+', 'concept\_5': '+'}\\
    Concepts contributions for Sample\_3: {'concept\_1': '+', 'concept\_5': '+', 'concept\_17': '+'}\\
    Concepts contributions for Sample\_4: {'concept\_11': '+'}\\
    Concepts contributions for Sample\_5: {'concept\_8': '+', 'concept\_10': '+', 'concept\_13': '+', 'concept\_19': '+'}\\
    Concepts contributions for Sample\_6: {'concept\_12': '+', 'concept\_14': '+'}\\
    Concepts contributions for Sample\_7: {'concept\_8': '+'}\\
    Concepts contributions for Sample\_8: {'concept\_3': '+', 'concept\_10': '+', 'concept\_11': '+'}\\
    Concepts contributions for Sample\_9: {'concept\_3': '-', 'concept\_9': '+', 'concept\_10': '+', 'concept\_13': '+', 'concept\_15': '+', 'concept\_19': '+'}
    \end{tcolorbox}
    \end{figure*}

    % --------------------
    % LP: Predictions
    \begin{figure*}[t]
    \begin{tcolorbox}[colframe=black!60,colback=white,title=2.3 LP: Predictions | \textit{NE2, E2, and E3}] 
    Sample\_0: physician\\
    Sample\_1: surgeon\\
    Sample\_2: nurse\\
    Sample\_3: physician\\
    Sample\_4: journalist\\
    Sample\_5: professor\\
    Sample\_6: dentist\\
    Sample\_7: psychologist\\
    Sample\_8: teacher\\
    Sample\_9: professor
    \end{tcolorbox}
    \end{figure*}

    % -----------
    % EP: Samples
    \begin{figure*}[t]
    \begin{tcolorbox}[colframe=black!60,colback=white,title=3.1 EP: Samples | \textit{In all prompt types}]
    Sample\_20:  He completed a master's degree in psychology at Duquesne University...\\
    Sample\_21:  Dr Shaileshwar Kumar practices at Oro Care Dental Clinic in Rajdanga...\\
    Sample\_22:  She has covered Kosovo, Israel, Palestine and Syria. Places of peril...\\
    Sample\_23:  He is the editor of The Costs of War: America\u2019s Pyrrhic...\\
    Sample\_24:  He currently practices at Buffalo Minimally Invasive Weight Loss...\\
    Sample\_25:  He accepts Coventry, United Healthcare HSA, United Healthcare HMO...\\ 
    Sample\_26:  He makes contact with Oskar Schindler, who astounds him with details...\\
    Sample\_27:  Davis of the Wyoming Supreme Court. Before joining the Supreme Court...\\
    Sample\_28:  She created her blog, Awash with Color, to inspire others, share tips...\\
    Sample\_29:  He did his undergraduate training from the esteemed Kathmandu...
    \end{tcolorbox}
    \end{figure*}

    % ------------------------------
    % Responses: user-LLM prediction
    \begin{figure*}[t]
    \begin{tcolorbox}[colframe=black!60,colback=white,title=3.2 Responses: user-LLM prediction | \textit{In all prompt types}]
    Sample\_20: psychologist\\
    Sample\_21: dentist\\
    Sample\_22: journalist\\
    Sample\_23: professor\\
    Sample\_24: physician\\
    Sample\_25: physician\\
    Sample\_26: journalist\\
    Sample\_27: attorney\\
    Sample\_28: teacher\\
    Sample\_29: professor
    \end{tcolorbox}
    \end{figure*}

\end{document}